\documentclass[10pt,twocolumn,letterpaper]{article}
\usepackage[pagenumbers]{cvpr}
\usepackage[T1]{fontenc}
\usepackage{multirow}
\usepackage{needspace}
\definecolor{datasetgreen}{RGB}{0,125,65}
\definecolor{datasetred}{RGB}{190,35,35}
\newcommand{\datayes}{\textcolor{datasetgreen}{\ensuremath{\checkmark}}}
\newcommand{\datano}{\textcolor{datasetred}{\ensuremath{\times}}}

\usepackage{tabularx}
\definecolor{cvprblue}{rgb}{0.21,0.49,0.74}
\usepackage[pagebackref,breaklinks,colorlinks,allcolors=cvprblue]{hyperref}
\hypersetup{pdfauthor={Jun Xu, Sujing Li, Chunhao Bi, Zhengxue Cheng, Ronghua Wu, Qunshan Gu, Li Song},pdftitle={SynAVID: A Synthetic Audio-Visual Immersive Dataset of Dynamic Scenes}}
\def\confName{CVPR}
\def\confYear{2027}
\title{SynAVID: A Synthetic Audio-Visual Immersive Dataset of Dynamic Scenes}
\author{\normalsize Jun Xu\textsuperscript{1}\quad Sujing Li\textsuperscript{1}\quad Chunhao Bi\textsuperscript{1}\quad Zhengxue Cheng\textsuperscript{1}\quad Ronghua Wu\textsuperscript{2}\quad Qunshan Gu\textsuperscript{2}\quad Li Song\textsuperscript{1}\\[3pt]
\textsuperscript{1}Shanghai Jiao Tong University\qquad \textsuperscript{2}Ant Group}

\begin{document}
\twocolumn[{
  \renewcommand\twocolumn[1][]{#1}
  \maketitle
  \begin{center}
    \includegraphics[trim=2mm 4mm 2mm 3mm, clip, width=\textwidth]{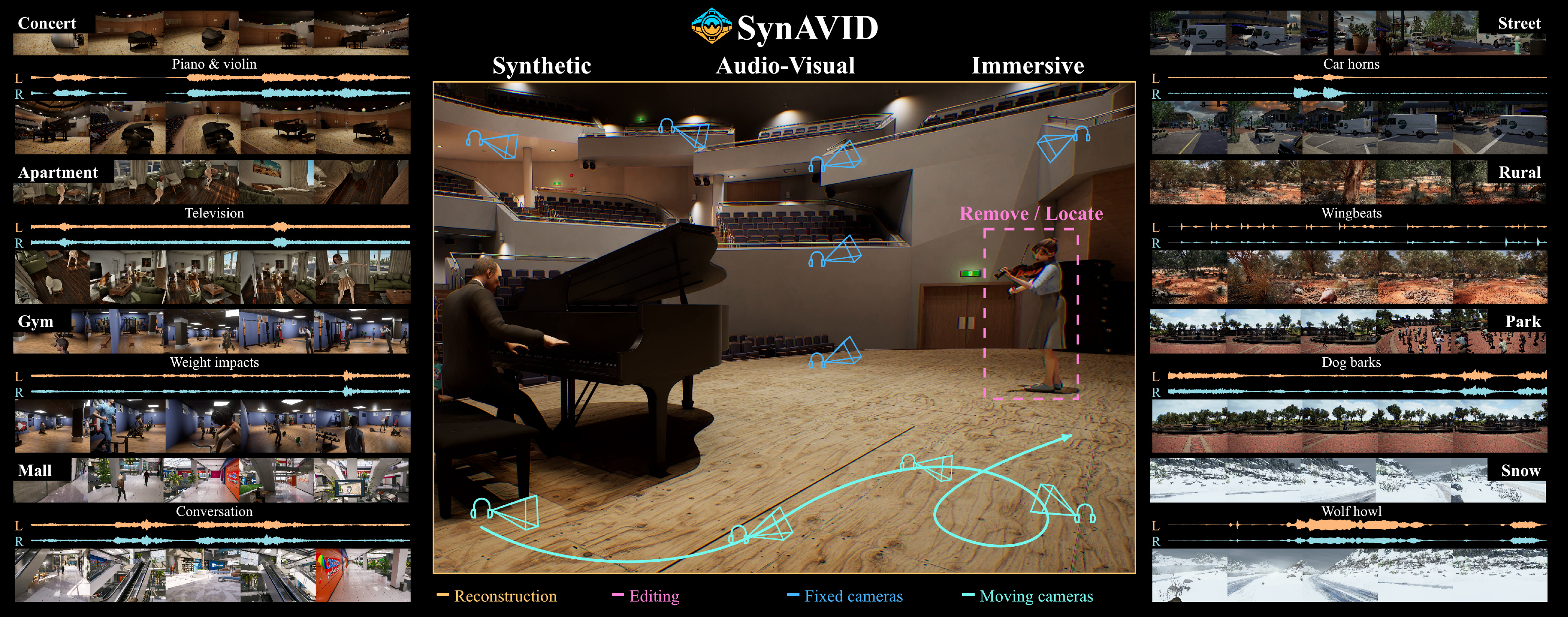}
    \vspace{-7mm}
    \captionof{figure}{\textbf{SynAVID overview.} Side panels show multiview images and selected binaural (L/R) waveforms from eight scenes. The center illustrates fixed and moving camera--listener pairs for audiovisual reconstruction and editing; placements and trajectories are schematic.}
    \label{fig:teaser}
    \vspace{-2mm}
  \end{center}
}]
\begingroup
\clubpenalty=10000
\widowpenalty=10000
\displaywidowpenalty=10000
\finalhyphendemerits=1000000
\begin{abstract}
Immersive scene modeling requires synchronized visual and auditory observations under changing viewpoints and scene conditions.
Obtaining such observations and references for controlled scene changes remains challenging.
We introduce \textbf{SynAVID}, a synthetic audio-visual immersive dataset comprising eight indoor and outdoor scenes.
Each minute-scale episode combines fixed camera arrays and moving trajectories with synchronized, viewpoint-dependent binaural audio.
Shared scene dynamics, sound-source events, and observer poses drive visual rendering and geometry-aware acoustic synthesis.
We evaluate existing methods for visual reconstruction, binaural audio reconstruction, and audiovisual editing.
Mixed-observation configurations generally improve visual reconstruction on fixed and moving test views, while transiently occupied and sparsely observed regions remain challenging.
Methods rank differently on spectral, interaural, and perceptual audio metrics.
Geometric and instance annotations expose uneven observation support and local Gaussian representations beyond aggregate scores.
Paired references assess target edits and surrounding content, while controlled geometry changes produce listener-dependent acoustic responses.
The project page is available at \href{https://avery-xu.github.io/synavid-project/}{\mbox{\textup{https://avery-xu.github.io/synavid-project/}}}.
\end{abstract}

\section{Introduction}
\label{sec:introduction}

\begin{table*}[t]
\centering
\caption{\textbf{Comparison with representative datasets.}
View counts and durations summarize reported capture configurations.
\textbf{Dynamic} refers to scene motion; \textbf{Fixed}/\textbf{Moving} to cameras. \textbf{Audio} includes recordings or impulse responses.
\textbf{GeoGT}: reference depth/surfaces, excluding reconstruction estimates; \textbf{Paired Edit}: matched original/edited captures.
\datayes/\datano: included/not included; ---: not applicable.}
\vspace{-3mm}
\label{tab:datasets}
\fontsize{8}{9.4}\selectfont
\setlength{\tabcolsep}{1.6pt}
\renewcommand{\arraystretch}{1.00}
\setlength{\aboverulesep}{1.5pt}
\setlength{\belowrulesep}{1.5pt}
\begin{tabular*}{\textwidth}{@{\extracolsep{\fill}}clcccccccccc@{}}
\toprule
 & \textbf{Dataset} & \textbf{Origin} & \textbf{Views} & \textbf{Duration} & \textbf{Dynamic} & \textbf{Fixed} & \textbf{Moving} & \textbf{Audio} & \textbf{Binaural} & \textbf{GeoGT} & \textbf{Paired Edit} \\
\midrule
\multirow{8}{*}{\rotatebox[origin=c]{90}{\fontsize{7.5}{9}\selectfont\textbf{Dynamic visual}}} & HyperNeRF~\cite{park2021hypernerf} & Real & 1--2 & 30--60\,s & \datayes & \datano & \datayes & \datano & \datano  & \datano & \datano \\
 & Neural 3D Video~\cite{li2022neural3dvideo} & Real & 18--21 & 10--40\,s & \datayes & \datayes & \datano & \datano & \datano  & \datano & \datano \\
 & Google Immersive~\cite{broxton2020immersive} & Real & 41--46 & 0.3--30\,s & \datayes & \datayes & \datano & \datano & \datano  & \datano & \datano \\
 & DiVa-360~\cite{lu2024diva360} & Real & 53 & 5--200\,s & \datayes & \datayes & \datano & \datayes & \datano  & \datano & \datano \\
 & Charge~\cite{nazarczuk2025charge} & Synth. & 41 & $\approx$1--7\,s & \datayes & \datayes & \datayes & \datano & \datano  & \datayes & \datano \\
 & Syn4D~\cite{jiang2026syn4d} & Synth. & 8 & $\approx 17$\,s & \datayes & \datayes & \datayes & \datano & \datano  & \datayes & \datano \\
 & MultiCamVideo~\cite{bai2025recammaster} & Synth. & 10 & 5.4\,s & \datayes & \datayes & \datayes & \datano & \datano  & \datano & \datano \\
 & StaDy4D~\cite{tsui2026stady4d} & Synth. & 9 & 5--30\,s & \datayes & \datayes & \datayes & \datano & \datano  & \datayes & \datayes \\
\midrule
\multirow{6}{*}{\rotatebox[origin=c]{90}{\fontsize{7.5}{9}\selectfont\textbf{Spatial acoustic}}} & RAF~\cite{chen2024raf} & Real & Images & --- & \datano & --- & --- & \datayes & \datano  & \datano & \datano \\
 & RWAVS~\cite{liang2023avnerf} & Real & 1 & 10--25\,min & \datano & \datano & \datayes & \datayes & \datayes  & \datano & \datano \\
 & MRTD~\cite{goetz2024mrtd} & Real & 360$^{\circ}$ images & --- & \datano & --- & --- & \datayes & \datayes  & \datano & \datano \\
 & AcousticRooms~\cite{liu2025acousticrooms} & Synth. & Depth maps & --- & \datano & --- & --- & \datayes & \datano  & \datayes & \datano \\
 & SoundSpaces-PanoIR~\cite{chen2022soundspaces2} & Synth. & RGB-D panos & --- & \datano & --- & --- & \datayes & \datano  & \datayes & \datano \\
 & SoundSpaces-NVAS~\cite{chen2023nvas} & Synth. & 4 images & --- & \datano & --- & --- & \datayes & \datayes  & \datayes & \datano \\
\midrule
\multirow{8}{*}{\rotatebox[origin=c]{90}{\fontsize{7.5}{9}\selectfont\textbf{Immersive audio-visual}}} & FAIR-Play~\cite{gao2019fairplay} & Real & 1 & 10\,s & \datayes & \datayes & \datano & \datayes & \datayes  & \datano & \datano \\
 & Tragic Talkers~\cite{berghi2022tragictalkers} & Real & 22 & $\approx$13--43\,s & \datayes & \datayes & \datano & \datayes & \datano  & \datano & \datano \\
 & EasyCom~\cite{donley2021easycom} & Real & 1 & 60\,s & \datayes & \datano & \datayes & \datayes & \datayes  & \datano & \datano \\
 & ImViD~\cite{yang2025imvid} & Real & 39 & 1--5\,min & \datayes & \datayes & \datayes & \datayes & \datano  & \datano & \datano \\
 & Replay-NVAS~\cite{chen2023nvas} & Real & 8 & 3--8\,min & \datayes & \datayes & \datano & \datayes & \datayes  & \datano & \datano \\
 & Ego-Exo4D~\cite{grauman2024egoexo4d} & Real & 5--6 & 8\,s--42\,min & \datayes & \datayes & \datayes & \datayes & \datano  & \datano & \datano \\
 & MRSAudio~\cite{guo2025mrsaudio} & Real & 1--2 & Mostly $<10$\,s & \datayes & \datayes & \datano & \datayes & \datayes  & \datano & \datano \\
 & SonoScene4D~\cite{jin2026sonoworld4d} & Real & 1 & 8\,s & \datayes & \datayes & \datano & \datayes & \datano  & \datano & \datano \\
\midrule
 & \textbf{SynAVID (ours)} & Synth. & \textbf{60} & \textbf{120\,s} & \datayes & \datayes & \datayes & \datayes & \datayes  & \datayes & \datayes \\
\bottomrule
\end{tabular*}
\vspace{-3mm}
\end{table*}

Creating digital environments that people can freely explore is a central goal of immersive computing, with applications in virtual reality, telepresence, and interactive media~\cite{mpeg2022immersivevideo}.
Such experiences depend on a coherent relationship between what users see and hear as they move through an evolving scene~\cite{itu2020immersiveav}.
Vision reveals appearance, motion, and occlusion, while spatial sound conveys the direction and activity of sources, including those outside the field of view, as well as their interaction with the surrounding environment.
Both modalities change with the observer's position and orientation and with the motion of people and objects~\cite{mpeg2024immersiveaudio}.
Modeling these relationships is essential for experiencing dynamic scenes from new perspectives with consistent visual and acoustic cues.
Developing and assessing such capabilities requires data that capture scene dynamics and observer-dependent visual and auditory responses within a common spatial and temporal frame.

Existing datasets have advanced dynamic visual reconstruction and spatial acoustic modeling along complementary directions.
Real and synthetic multiview datasets provide rich observations of dynamic appearance and geometry, with synthetic data additionally offering dense geometric annotations, but their primary emphasis remains visual reconstruction and understanding~\cite{li2022neural3dvideo,nazarczuk2025charge,jiang2026syn4d}.
Acoustic datasets and simulators characterize how sound varies with source and receiver configurations, but commonly focus on static scene geometry~\cite{chen2024raf,liang2023avnerf,chen2022soundspaces2}.
Dynamic audiovisual recordings connect these directions by capturing visual and auditory observations of evolving scenes~\cite{yang2025imvid,chen2023nvas}.
However, physical acquisition offers limited independent control over scene evolution and observation conditions, making it difficult to systematically study their effects on visual and acoustic reconstruction.
Together, these directions motivate data that jointly capture dynamic visual and acoustic behavior while allowing systematic control over scene content and observation conditions.

We introduce \textbf{SynAVID}, a synthetic audio-visual immersive dataset spanning dynamic indoor and outdoor environments (Figure~\ref{fig:teaser}).
SynAVID captures minute-scale activities using dense fixed camera arrays together with moving trajectories, providing complementary observations of the same evolving scene.
Each camera is paired with a colocated binaural listener, producing synchronized video and spatial audio corresponding to the observer's position and orientation.
A shared scene configuration and event timeline drive visual rendering and geometry-aware acoustic synthesis, linking visible motion, sound-source activity, and changes in sound propagation.
This shared configuration allows observation trajectories to vary independently of scene activity, while geometric annotations and paired editing references support regional reconstruction diagnostics and assessment of target edits.

Our contributions are threefold:
\begin{itemize}
    \item SynAVID, a synthetic dataset of dynamic indoor and outdoor scenes with fixed and moving multiview video and synchronized, listener-dependent binaural audio.
    \item A generation workflow that replays shared scene and sound-event timelines along alternative camera--listener trajectories and produces reference observations for controlled scene edits.
    \item Evaluations of visual reconstruction, binaural audio reconstruction, and audiovisual editing using standard metrics, complemented by regional diagnostics and a controlled dynamic-occlusion case.
\end{itemize}

\section{Related Work}
\label{sec:related}

\begin{figure*}[t]
    \centering
    \includegraphics[width=\textwidth]{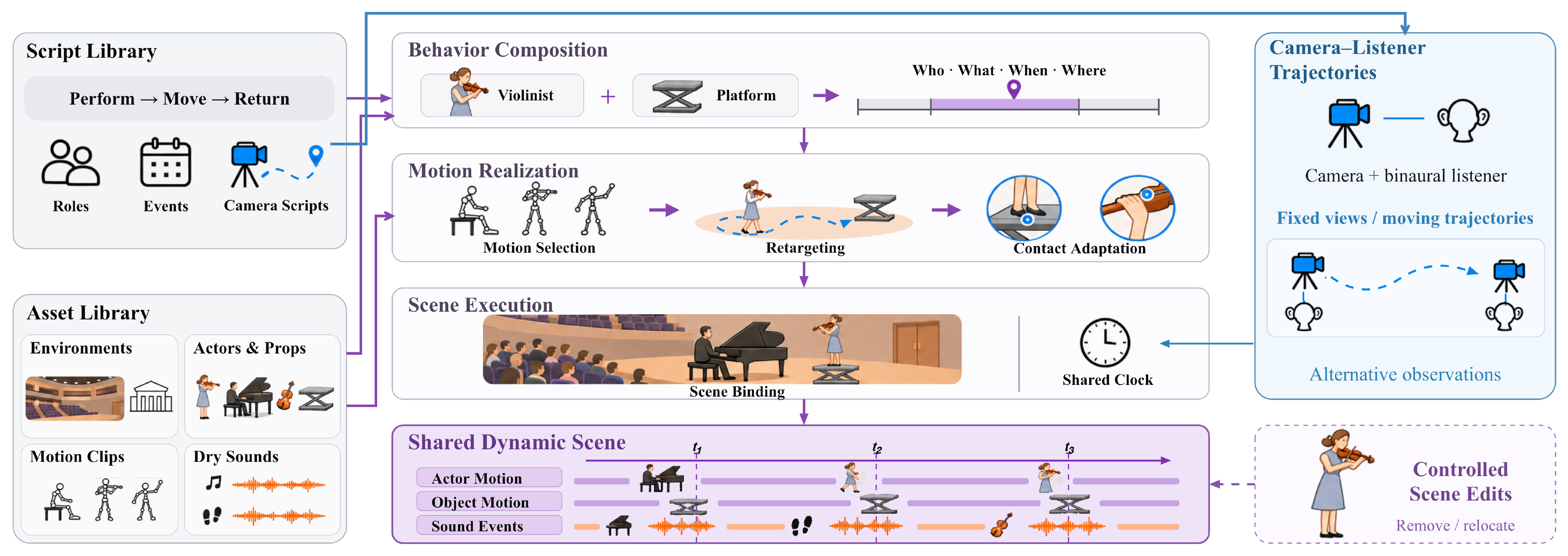}
    \vspace{-7mm}\caption{\textbf{Scene organization in SynAVID.} Asset and script libraries support behavior composition, motion realization, and scene execution, producing actor-motion, object-motion, and sound-event tracks on a shared timeline. Camera--listener trajectories specify fixed or moving observations of the same episode. The dashed intervention denotes controlled changes to selected scene elements; observation trajectories and scene content can be varied separately.}
    
\label{fig:scene_organization}
\vspace{-5mm}
\end{figure*}

Table~\ref{tab:datasets} compares representative datasets across visual, acoustic, and joint audiovisual observation.

\noindent\textbf{Dynamic visual datasets.}
Datasets commonly used in dynamic Gaussian reconstruction include D-NeRF~\cite{pumarola2021dnerf} and HyperNeRF~\cite{park2021hypernerf} for monocular settings.
Neural 3D Video (N3V)~\cite{li2022neural3dvideo}, Google Immersive~\cite{broxton2020immersive}, Technicolor~\cite{sabater2017lightfield}, and DiVa-360~\cite{lu2024diva360} provide complementary synchronized multiview recordings of dynamic scenes.
Synthetic datasets extend control over scene content and observation conditions: Charge~\cite{nazarczuk2025charge} and Syn4D~\cite{jiang2026syn4d} combine varied camera configurations with geometric supervision, while MultiCamVideo~\cite{bai2025recammaster} emphasizes diverse synchronized camera trajectories.
StaDy4D~\cite{tsui2026stady4d} additionally provides paired captures with and without dynamic actors.
These resources support visual reconstruction through complementary capture layouts, scene dynamics, and reference annotations.

\noindent\textbf{Spatial acoustic datasets and simulation.}
Measured resources range from visually aligned monaural room responses in RAF~\cite{chen2024raf} to moving-listener binaural recordings paired with images in RWAVS~\cite{liang2023avnerf}.
Simulation complements these measurements with controllable source, receiver, and environment configurations.
SoundSpaces~2.0~\cite{chen2022soundspaces2} enables continuous acoustic queries, and SoundSpaces-NVAS~\cite{chen2023nvas} pairs rendered views with binaural speech.
AcousticRooms~\cite{liu2025acousticrooms} varies room geometry and materials, while AudioWorldSim~\cite{zerkowski2026audioworldsim} generates binaural navigation sequences.
These resources primarily characterize acoustic variation within static scene geometry, even when signals or listeners change over time.

\noindent\textbf{Immersive audio-visual datasets.}
Joint recordings capture dynamic activities with complementary visual and acoustic observations.
ImViD~\cite{yang2025imvid} provides minute-scale multiview video with camera audio, while Replay-NVAS~\cite{chen2023nvas} supplies synchronized views with measured binaural targets.
Ego-Exo4D~\cite{grauman2024egoexo4d} combines moving egocentric and fixed exocentric views with device audio.
MRSAudio~\cite{guo2025mrsaudio} includes video paired with binaural and Ambisonic recordings across its subsets; SonoScene4D~\cite{jin2026sonoworld4d} pairs panoramic video with first-order Ambisonic recordings at a separate listening position.
Their references are tied to recorded configurations, limiting controlled comparisons after scene changes.
SynAVID complements these resources with a shared scene and timeline coordinating dynamic geometry, source events, and fixed and moving camera--listener pairs.
Its synchronized video and modeled binaural references support joint evaluation under known conditions and controlled scene changes.

\section{Dataset and Production Pipeline}
\label{sec:dataset}

SynAVID separates the construction of a dynamic episode (Figure~\ref{fig:scene_organization}) from its observation through paired cameras and listeners (Figure~\ref{fig:production_pipeline}).

\begin{figure*}[!t]
    \centering
    \includegraphics[width=\textwidth]{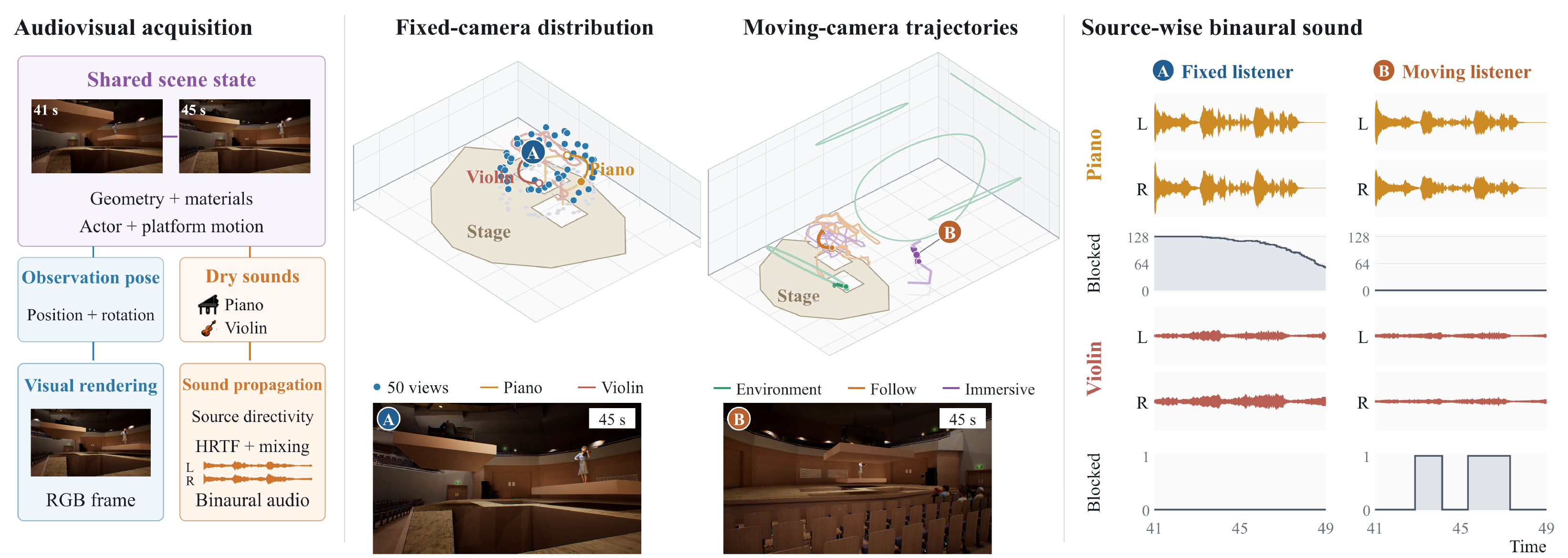}
    \vspace{-7mm}
\caption{\textbf{Audiovisual acquisition and example observations.} Shared scene states drive RGB rendering and spatial sound synthesis (left). Fixed cameras, moving paths, and source motion are shown relative to the Concert stage at different zoom levels (middle). Pale paths span the episode; darker segments highlight 41--49\,s on featured trajectories. Green/orange/purple distinguish environment/follow/immersive cameras. A and B pair RGB at 45\,s with source-specific L/R sound over 41--49\,s (right). Mirrored RMS envelopes show total received sound on a shared amplitude scale; \textit{Blocked} counts occluded visibility samples. Modeling details are in the supplement.}
    
\label{fig:production_pipeline}
\vspace{-5mm}
\end{figure*}

\subsection{Scene Organization}
\label{sec:dataset_organization}

Scene content, scripted behavior, and observation trajectories share a common clock.
An episode defines the environment, actors, objects, and scheduled sound sources; independently specified camera--listener trajectories determine how it is seen and heard.

Figure~\ref{fig:scene_organization} organizes reusable assets and scripts into three stages.
\emph{Behavior composition} assigns participants, actions, and sound events in space and time.
\emph{Motion realization} selects and adapts motion clips and contacts.
\emph{Scene execution} binds motions and emitters to scene instances and evaluates their states and source activity along the timeline.

Alternative camera--listener trajectories observe an unchanged episode.
Controlled edits modify selected elements for comparison with the original configuration.

\subsection{Dynamic Scene Construction}
\label{sec:dataset_construction}

Within each episode, sustained activities combine with local movements and interactions, changing both visibility and sound propagation.

Motion clips are retargeted to the characters and connected into longer performances with short blends.
Routes place these actions in the environment; local inverse kinematics and contact adjustments align feet, hands, and held objects with supporting surfaces.
Object motion follows the same timeline, including platforms that carry performers.

Dry signals are bound to actors, objects, or fixed emitters, with scheduled segments and gains defining what is emitted and when.
Emitter trajectories locate each sound; footsteps and contact sounds follow actions, while music and speech occupy assigned intervals.
Musical playback is coordinated at the event level.

\subsection{Synchronized Audio-Visual Acquisition}
\label{sec:dataset_acquisition}

\noindent\textbf{Complementary viewpoints.}
The shared scene state in Figure~\ref{fig:production_pipeline} (left) supplies the visual and acoustic inputs for each camera--listener pair.
Fixed arrays repeatedly observe activity regions from multiple directions.
Environment, follow, and immersive trajectories add changing viewing distances, angles, and scene coverage.
They bring additional surfaces into view for varying durations, creating both persistent and intermittent observations within one episode.
Figure~\ref{fig:production_pipeline} (middle) shows a fixed array around the Concert stage and moving paths extending through the surrounding space, alongside the changing positions of the piano and violin sources.
Unreal Engine renders native 4K RGB at 30\,fps with corresponding camera intrinsics and poses.

\noindent\textbf{Listener-dependent sound.}
Source and listener poses, directivity, and intervening geometry shape the sound received at each viewpoint.
Steam Audio~\cite{valveSteamAudio} models direct and reflected propagation through successive quasi-static scene states.
Acoustic materials draw on AcousticRooms~\cite{liu2025acousticrooms} and presets; the piano uses a measured directivity model~\cite{brandner2020piano}.
In Figure~\ref{fig:production_pipeline} (right), listener A receives piano sound despite strong direct-path occlusion, while listener B encounters intervals of violin occlusion.
These total-sound envelopes include reflections, so direct-path blockage need not produce silence.
Source-specific waveforms reveal how each sound contributes to the listener's mixture.

\begin{figure*}[!t]
    \centering
    \includegraphics[width=\textwidth]{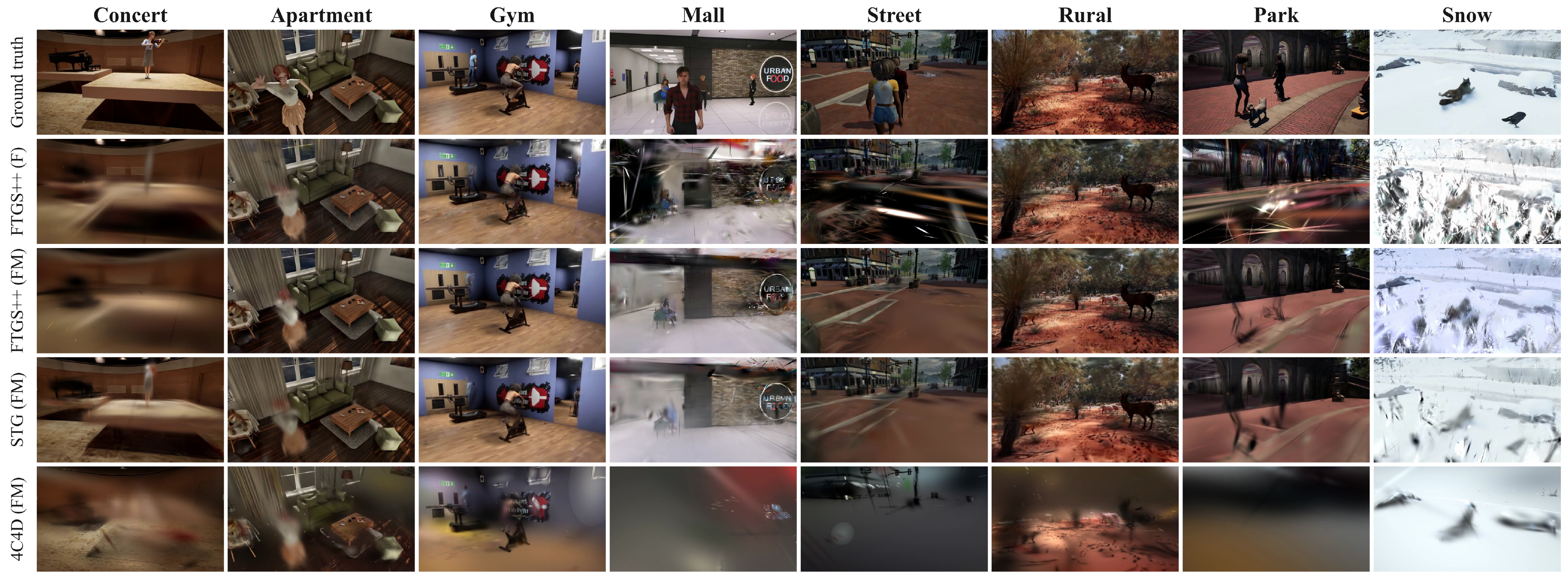}
    \vspace{-7mm}\caption{\textbf{Novel-view reconstruction across scenes.}
    Columns show scenes and rows show GT and methods, with matched test cameras and times within each column.
    F/FM denote fixed/mixed training views.
    Frames are decoded from compressed videos; Table~\ref{tab:visual_roaming} uses floating-point renders.}
    \vspace{-5mm}\label{fig:visual_reconstruction}
\end{figure*}

\noindent\textbf{Paired audiovisual observations.}
A listener shares each camera's position and orientation.
The FABIAN HRTF~\cite{brinkmann2017fabian} maps source responses to the listener's ears, yielding the mixture at view $c$ and ear $e\in\{L,R\}$:
\begin{equation}
    y_c^e[n] = \sum_s g_s \sum_k h_{s,c}^e[n,k]x_s[n-k],
    \label{eq:binaural_mixture}
\end{equation}
Here $x_s$ is the scheduled dry signal, $h_{s,c}^e$ the time-varying binaural filter, and $g_s$ a source gain shared across views.
Images and 48\,kHz audio correspond to the same scene events, while received sound retains propagation delays, reverberation tails, and relative levels across listeners.
Implementation details are in the supplement.

\begin{table}[t]
    \centering
    \caption{\textbf{Scene coverage.} Representative dynamics and sound sources in four indoor (top) and four outdoor (bottom) environments. Each 120-second episode combines 50 fixed views and 10 moving camera--listener trajectories.}
    \vspace{-3mm}\label{tab:scene_content}
    \small
    \setlength{\tabcolsep}{3pt}
    \renewcommand{\arraystretch}{1.12}
    \begin{tabular*}{\columnwidth}{@{\extracolsep{\fill}}lll@{}}
        \toprule
        \textbf{Scene} & \textbf{Dynamics} & \textbf{Main sounds} \\
        \midrule
        Concert & Performance, platforms & Piano, violin \\
        Apartment & Dance, furniture motion & TV, dog, robot \\
        Gym & Exercise, equipment use & Music, speech, impacts \\
        Mall & Walking, escalators & Speech, footsteps \\
        \midrule
        Street & Pedestrians, traffic & Engines, horns, speech \\
        Rural & Animals, vehicle & Calls, vehicle \\
        Park & Exercise, social activity & Music, speech, dogs \\
        Snow & Animals, vehicle & Calls, snow contact, vehicle \\
        \bottomrule
    \end{tabular*}
    \vspace{-6mm}
\end{table}

\subsection{Dataset Overview}
\label{sec:dataset_products}

SynAVID covers four indoor and four outdoor environments (Table~\ref{tab:scene_content}).
Each contains one 120-second episode observed from 50 fixed cameras and 10 moving trajectories: 16 minutes of distinct scene time, 480 view sequences, and 16 view-hours in total.

For each view, the base release provides RGB video, a mixed binaural waveform, camera and listener parameters, timestamps, and split metadata.
Source data include dry signals, emission timelines, and emitter bindings.

The same episodes support additional trajectories and the export of depth, surface labels, and geometry on request.
The supplement details release contents, reference coverage, acoustic settings, and implementation checks.

\par\endgroup
\section{Experiments}
\label{sec:experiments}

\begin{table*}[!tp]
    \centering
    \caption{\textbf{Visual reconstruction on fixed and moving test cameras.}
    F: 46 fixed training views; FM: 53 indoor or 54 outdoor fixed and moving training views.
    Scores average frames within each test route, then routes within the fixed or moving group.
    \textbf{Bold}/\underline{underline}: best/second per scene and metric.
    FTGS++ F/FM also differ in training-derived initialization; full settings appear in the supplement.}
    \vspace{-3mm}\label{tab:visual_roaming}
    \footnotesize
    \setlength{\tabcolsep}{1.5pt}
    \renewcommand{\arraystretch}{1.02}
    \setlength{\aboverulesep}{1.5pt}
    \setlength{\belowrulesep}{1.5pt}
    \begin{tabular*}{\textwidth}{@{\extracolsep{\fill}}l*{6}{c}@{\hspace{7pt}}*{6}{c}@{}}
        \toprule
        \multirow{2}{*}{Method} & \multicolumn{3}{c}{Fixed test} & \multicolumn{3}{c}{Moving test} & \multicolumn{3}{c}{Fixed test} & \multicolumn{3}{c}{Moving test} \\
        \cmidrule(lr){2-4}\cmidrule(lr){5-7}\cmidrule(lr){8-10}\cmidrule(lr){11-13}
        & PSNR$\uparrow$ & SSIM$\uparrow$ & LPIPS$\downarrow$ & PSNR$\uparrow$ & SSIM$\uparrow$ & LPIPS$\downarrow$ & PSNR$\uparrow$ & SSIM$\uparrow$ & LPIPS$\downarrow$ & PSNR$\uparrow$ & SSIM$\uparrow$ & LPIPS$\downarrow$ \\
        \midrule
         & \multicolumn{6}{c}{\textbf{Concert}} & \multicolumn{6}{c}{\textbf{Apartment}} \\
        \cmidrule(lr){2-7}\cmidrule(lr){8-13}
        FTGS++$^{\mathrm{F}}$~\cite{lee2026freetimegspp} & \underline{22.422} & 0.6218 & 0.5414 & 16.868 & 0.5306 & 0.6153 & 20.570 & 0.7810 & \underline{0.1888} & 21.422 & 0.7798 & \underline{0.1742} \\
        FTGS++$^{\mathrm{FM}}$~\cite{lee2026freetimegspp} & 20.284 & 0.5691 & 0.6643 & 17.749 & 0.5866 & 0.6684 & \textbf{22.001} & \textbf{0.8351} & \textbf{0.1518} & \textbf{22.908} & \textbf{0.8466} & \textbf{0.1324} \\
        STG$^{\mathrm{FM}}$~\cite{li2024spacetime} & \textbf{25.443} & \textbf{0.7775} & \textbf{0.2641} & \textbf{19.889} & \textbf{0.6587} & \textbf{0.3821} & \underline{21.462} & \underline{0.8024} & 0.2028 & \underline{22.178} & \underline{0.8080} & 0.1810 \\
        4C4D$^{\mathrm{FM}}$~\cite{zhou20264c4d} & 20.864 & \underline{0.6812} & \underline{0.4119} & \underline{18.478} & \underline{0.6178} & \underline{0.5218} & 17.884 & 0.6239 & 0.4722 & 18.753 & 0.6303 & 0.4492 \\
        \midrule
         & \multicolumn{6}{c}{\textbf{Gym}} & \multicolumn{6}{c}{\textbf{Mall}} \\
        \cmidrule(lr){2-7}\cmidrule(lr){8-13}
        FTGS++$^{\mathrm{F}}$~\cite{lee2026freetimegspp} & 19.877 & 0.7503 & 0.2261 & 21.858 & 0.7917 & \underline{0.1883} & 13.012 & 0.3555 & 0.5071 & 12.860 & 0.3812 & 0.5027 \\
        FTGS++$^{\mathrm{FM}}$~\cite{lee2026freetimegspp} & \textbf{23.346} & \textbf{0.8553} & \textbf{0.1329} & \textbf{25.403} & \textbf{0.8869} & \textbf{0.1163} & \textbf{18.023} & \textbf{0.6337} & \textbf{0.3029} & \textbf{17.629} & \textbf{0.6770} & \textbf{0.2849} \\
        STG$^{\mathrm{FM}}$~\cite{li2024spacetime} & \underline{22.713} & \underline{0.8240} & \underline{0.2134} & \underline{24.018} & \underline{0.8404} & 0.2069 & \underline{17.572} & \underline{0.5994} & \underline{0.3846} & \underline{16.899} & \underline{0.6269} & \underline{0.3791} \\
        4C4D$^{\mathrm{FM}}$~\cite{zhou20264c4d} & 15.030 & 0.6235 & 0.5720 & 18.190 & 0.6976 & 0.4929 & 11.788 & 0.4106 & 0.8021 & 11.631 & 0.4368 & 0.8133 \\
        \midrule
         & \multicolumn{6}{c}{\textbf{Street}} & \multicolumn{6}{c}{\textbf{Rural}} \\
        \cmidrule(lr){2-7}\cmidrule(lr){8-13}
        FTGS++$^{\mathrm{F}}$~\cite{lee2026freetimegspp} & \underline{25.151} & \underline{0.8082} & \underline{0.1685} & 14.874 & 0.3555 & 0.5376 & 17.949 & 0.4686 & \underline{0.3468} & 16.710 & 0.3636 & \underline{0.4683} \\
        FTGS++$^{\mathrm{FM}}$~\cite{lee2026freetimegspp} & \textbf{26.792} & \textbf{0.8501} & \textbf{0.1377} & \textbf{21.823} & \textbf{0.7308} & \textbf{0.3115} & \textbf{19.905} & \textbf{0.5584} & \textbf{0.2999} & \textbf{18.841} & \textbf{0.4698} & \textbf{0.4363} \\
        STG$^{\mathrm{FM}}$~\cite{li2024spacetime} & 24.973 & 0.7907 & 0.2505 & \underline{20.532} & \underline{0.6663} & \underline{0.4596} & \underline{19.267} & \underline{0.4978} & 0.4320 & \underline{17.818} & \underline{0.4113} & 0.5458 \\
        4C4D$^{\mathrm{FM}}$~\cite{zhou20264c4d} & 19.419 & 0.5868 & 0.5497 & 17.654 & 0.5442 & 0.7249 & 16.833 & 0.3110 & 0.7600 & 15.957 & 0.3104 & 0.7440 \\
        \midrule
         & \multicolumn{6}{c}{\textbf{Park}} & \multicolumn{6}{c}{\textbf{Snow}} \\
        \cmidrule(lr){2-7}\cmidrule(lr){8-13}
        FTGS++$^{\mathrm{F}}$~\cite{lee2026freetimegspp} & \underline{21.885} & 0.6144 & \underline{0.2551} & 15.577 & 0.3631 & \underline{0.5034} & 20.309 & 0.7914 & 0.2307 & 13.985 & 0.5714 & 0.4713 \\
        FTGS++$^{\mathrm{FM}}$~\cite{lee2026freetimegspp} & \textbf{22.132} & \textbf{0.6436} & \textbf{0.2468} & \textbf{19.407} & \textbf{0.5266} & \textbf{0.4068} & \underline{22.216} & \textbf{0.8649} & \textbf{0.1677} & \textbf{17.970} & \textbf{0.7626} & \textbf{0.2992} \\
        STG$^{\mathrm{FM}}$~\cite{li2024spacetime} & 21.092 & \underline{0.6402} & 0.3306 & \underline{18.745} & \underline{0.5002} & 0.5068 & \textbf{22.659} & \underline{0.8217} & \underline{0.2300} & \underline{17.864} & \underline{0.7408} & \underline{0.3241} \\
        4C4D$^{\mathrm{FM}}$~\cite{zhou20264c4d} & 15.643 & 0.4319 & 0.6635 & 15.158 & 0.3648 & 0.8225 & 18.100 & 0.6984 & 0.4192 & 14.775 & 0.6308 & 0.5234 \\
        \bottomrule
    \end{tabular*}
    \vspace{-1mm}
\end{table*}

\begin{figure*}[!t]
    \centering
    \includegraphics[width=\textwidth]{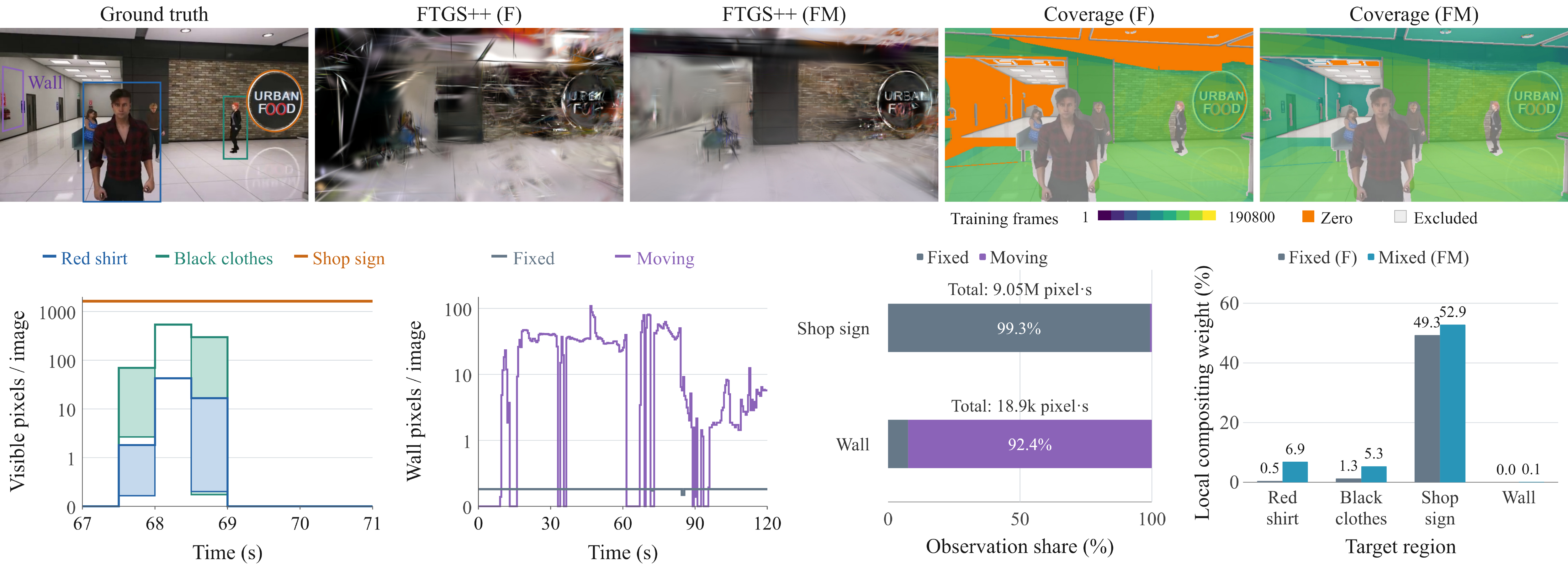}
    \vspace{-7mm}\caption{\textbf{Mall observation and representation diagnostics.}
    Top: a shared test view, FTGS++ reconstructions, and static-surface coverage (log scale; orange: zero; faded: excluded).
    Bottom, left to right: support at fixed person/sign locations, wall support from fixed and moving views, cumulative observation shares and pixel-second totals, and nearby Gaussians' mean compositing weight over target pixels.
    Support curves show pixels per training image on a symlog scale; shading denotes lower and upper geometric attribution bounds.
    Person support is zero at sampled times outside 67--71\,s; totals integrate support over views and time.}
    \vspace{-5mm}\label{fig:mall_observation_representation}
\end{figure*}

Our experiments cover dynamic visual reconstruction (Section~\ref{sec:exp_visual}), binaural audio reconstruction (Section~\ref{sec:exp_audio}), and audiovisual editing (Section~\ref{sec:exp_editing}).
Alongside reconstruction and editing results, we use regional annotations and a controlled geometric intervention to examine local visual representations and listener-dependent sound propagation.

\Needspace{7\baselineskip}
\subsection{Visual Reconstruction}
\label{sec:exp_visual}

We evaluate FreeTimeGS++ (FTGS++)~\cite{lee2026freetimegspp}, Spacetime Gaussians (STG)~\cite{li2024spacetime}, and 4C4D~\cite{zhou20264c4d} across all eight scenes.
Each method fits one model per 120-second sequence.
FTGS++ uses fixed-only (F) or mixed fixed and moving (FM) training views; STG and 4C4D use FM.
We evaluate on both fixed and moving held-out views.

\begin{table*}[!tp]
    \centering
    \caption{\textbf{Binaural audio reconstruction.}
    Errors at held-out fixed listener poses, using a synchronized mixture from another view as input.
    Binaural copy preserves the input L/R channels without spatial adaptation; DPAM/CDPAM use mono fold-downs.
    Lower is better. LRE: dB; RTE: s; MSE: $\times 10^3$. \textbf{Bold}/\underline{underline}: best/second per scene and metric.}
    \vspace{-3mm}\label{tab:audio_observed}
    \footnotesize
    \setlength{\tabcolsep}{1pt}
    \renewcommand{\arraystretch}{1.04}
    \setlength{\aboverulesep}{1.8pt}
    \setlength{\belowrulesep}{1.8pt}
    \begin{tabular*}{\textwidth}{@{\extracolsep{\fill}}l*{8}{c}@{\hspace{7pt}}*{8}{c}@{}}
        \toprule
        Method & MAG$\downarrow$ & ENV$\downarrow$ & LRE$\downarrow$ & RTE$\downarrow$ & STFT$\downarrow$ & MSE$\downarrow$ & DPAM$\downarrow$ & CDPAM$\downarrow$ & MAG$\downarrow$ & ENV$\downarrow$ & LRE$\downarrow$ & RTE$\downarrow$ & STFT$\downarrow$ & MSE$\downarrow$ & DPAM$\downarrow$ & CDPAM$\downarrow$ \\
        \midrule
         & \multicolumn{8}{c}{\textbf{Concert}} & \multicolumn{8}{c}{\textbf{Apartment}} \\
        \cmidrule(lr){2-9}\cmidrule(lr){10-17}
        AudioGS~\cite{bi2026audiogs} & 0.1447 & 0.06541 & 1.871 & 0.0337 & 206.15 & 3.200 & \textbf{3.6573} & \textbf{0.15566} & \underline{0.0918} & 0.04138 & 0.796 & \underline{0.0298} & \underline{148.14} & \underline{1.966} & \underline{3.6566} & \textbf{0.15483} \\
        ViGAS~\cite{chen2023nvas} & \textbf{0.1132} & \textbf{0.05708} & \textbf{1.008} & \textbf{0.0281} & \textbf{189.30} & \textbf{2.725} & 3.8276 & 0.66120 & \textbf{0.0772} & \textbf{0.03684} & \textbf{0.555} & \textbf{0.0272} & 148.68 & 1.989 & 3.7772 & 1.04410 \\
        AV-Cloud~\cite{chen2024avcloud} & \underline{0.1310} & \underline{0.06193} & \underline{1.246} & \underline{0.0288} & \underline{202.92} & \underline{3.046} & \underline{3.7004} & \underline{0.15615} & 0.0921 & \underline{0.03882} & \underline{0.698} & 0.0310 & \textbf{140.07} & \textbf{1.746} & 3.7567 & 0.25978 \\
        Binaural copy & 0.1634 & 0.07386 & 1.971 & 0.0358 & 224.69 & 3.747 & 3.7031 & 0.16021 & 0.1040 & 0.04383 & 0.949 & 0.0299 & 152.88 & 2.108 & \textbf{3.6552} & \underline{0.19700} \\
        \midrule
         & \multicolumn{8}{c}{\textbf{Gym}} & \multicolumn{8}{c}{\textbf{Mall}} \\
        \cmidrule(lr){2-9}\cmidrule(lr){10-17}
        AudioGS~\cite{bi2026audiogs} & 0.0732 & \underline{0.02380} & 0.743 & 0.0338 & \underline{71.61} & 0.484 & \textbf{3.7549} & \textbf{0.21796} & 0.1314 & 0.04377 & 1.449 & \textbf{0.0714} & 105.83 & 0.949 & \underline{3.9014} & \textbf{0.24507} \\
        ViGAS~\cite{chen2023nvas} & \textbf{0.0615} & \textbf{0.02180} & \underline{0.658} & \underline{0.0283} & \textbf{66.99} & \textbf{0.432} & 4.0117 & 1.25600 & \textbf{0.1055} & \textbf{0.03582} & \underline{1.165} & 0.0816 & \textbf{80.79} & \textbf{0.587} & 3.9555 & 1.22420 \\
        AV-Cloud~\cite{chen2024avcloud} & \underline{0.0667} & 0.02383 & \textbf{0.623} & \textbf{0.0268} & 71.85 & \underline{0.480} & 3.8734 & 0.23328 & \underline{0.1087} & \underline{0.03629} & \textbf{1.066} & 0.1027 & \underline{85.34} & \underline{0.628} & 4.0022 & 0.26779 \\
        Binaural copy & 0.0758 & 0.02444 & 0.771 & 0.0333 & 72.92 & 0.499 & \underline{3.8107} & \underline{0.23213} & 0.1338 & 0.04577 & 1.568 & \underline{0.0776} & 108.30 & 0.998 & \textbf{3.8639} & \underline{0.24631} \\
        \midrule
         & \multicolumn{8}{c}{\textbf{Street}} & \multicolumn{8}{c}{\textbf{Rural}} \\
        \cmidrule(lr){2-9}\cmidrule(lr){10-17}
        AudioGS~\cite{bi2026audiogs} & 0.0280 & 0.00675 & 2.359 & 0.0523 & 16.94 & 0.02177 & \textbf{3.9018} & \underline{0.38975} & 0.0222 & 0.00530 & 3.702 & \textbf{0.0432} & 12.34 & 0.02271 & \underline{3.6817} & \underline{0.19429} \\
        ViGAS~\cite{chen2023nvas} & \textbf{0.0199} & \textbf{0.00519} & \textbf{1.168} & \textbf{0.0382} & \textbf{13.36} & \textbf{0.01448} & 3.9517 & \textbf{0.22257} & \textbf{0.0182} & \textbf{0.00439} & \underline{2.872} & 0.0540 & \textbf{10.42} & \textbf{0.01566} & \textbf{3.6695} & \textbf{0.18888} \\
        AV-Cloud~\cite{chen2024avcloud} & \underline{0.0219} & \underline{0.00540} & \underline{1.419} & \underline{0.0439} & \underline{14.71} & \underline{0.01709} & 4.0737 & 0.44165 & \underline{0.0196} & \underline{0.00467} & \textbf{2.544} & 0.0632 & \underline{11.34} & \underline{0.01763} & 3.7695 & 0.26065 \\
        Binaural copy & 0.0312 & 0.00787 & 2.420 & 0.0561 & 18.83 & 0.02610 & \underline{3.9252} & 0.42361 & 0.0256 & 0.00607 & 3.724 & \underline{0.0435} & 13.77 & 0.02899 & 3.7129 & 0.19979 \\
        \midrule
         & \multicolumn{8}{c}{\textbf{Park}} & \multicolumn{8}{c}{\textbf{Snow}} \\
        \cmidrule(lr){2-9}\cmidrule(lr){10-17}
        AudioGS~\cite{bi2026audiogs} & 0.0490 & 0.01338 & 1.961 & 0.0613 & 32.73 & 0.09613 & \underline{3.8707} & 0.23156 & 0.0181 & 0.00743 & 4.897 & \textbf{0.0496} & 16.95 & 0.17798 & \textbf{3.7080} & \textbf{0.20177} \\
        ViGAS~\cite{chen2023nvas} & \textbf{0.0344} & \textbf{0.00913} & \underline{1.374} & \textbf{0.0498} & \textbf{23.49} & \textbf{0.05265} & \textbf{3.8645} & \textbf{0.14499} & \textbf{0.0121} & \textbf{0.00425} & \underline{3.805} & 0.0794 & \textbf{11.26} & \underline{0.06485} & 3.8747 & 0.22305 \\
        AV-Cloud~\cite{chen2024avcloud} & \underline{0.0388} & \underline{0.00970} & \textbf{1.270} & \underline{0.0601} & \underline{26.04} & \underline{0.06055} & 3.9648 & \underline{0.21445} & \underline{0.0157} & \underline{0.00483} & \textbf{3.550} & 0.0841 & \underline{11.91} & \textbf{0.06000} & 3.8664 & 0.37259 \\
        Binaural copy & 0.0635 & 0.01941 & 2.120 & 0.0629 & 41.62 & 0.14419 & 3.8834 & 0.26267 & 0.0211 & 0.00946 & 5.292 & \underline{0.0522} & 20.14 & 0.26047 & \underline{3.7585} & \underline{0.22046} \\
        \bottomrule
    \end{tabular*}
\end{table*}

Table~\ref{tab:visual_roaming} reports PSNR, SSIM, and LPIPS for both test groups.
FM FTGS++ improves over F across all metrics in seven scenes, with outdoor moving-view PSNR gains of 2.13--6.95\,dB.
Concert remains an exception: STG performs best, and FM does not consistently improve FTGS++.
In Figure~\ref{fig:visual_reconstruction}, FM reduces large background artifacts in Street and Park, while people remain missing or ghosted.
The selected reconstructions retain more recognizable subjects in Apartment, Gym, and Rural, whereas people in Concert and Mall are severely blurred or nearly absent.
These results show improved reconstruction under mixed observations alongside persistent local failures, which we examine more closely in Mall.

\noindent\textbf{Observation support.}
Prior work highlights temporally uneven supervision in dynamic reconstruction~\cite{sandu2026temporally}.
Figure~\ref{fig:mall_observation_representation} contrasts three support patterns in one Mall view at 68\,s: transiently occupied person locations, a persistently observed shop sign, and a corridor wall mainly covered by moving cameras.
At 2\,Hz, geometry and instance labels estimate visible pixels at these fixed world regions; summing over views and integrating over time gives pixel-seconds.
The corridor wall receives about 92\% of its projected-pixel support from moving views across multiple time intervals, but its cumulative total remains far below the sign's.
The upper row juxtaposes FM's improved wall appearance with missing people under these distinct support patterns.

\noindent\textbf{Final representation.}
Recent work also examines missing dynamic content through the allocation of Gaussians~\cite{roh2026dusting}.
We measure local representation by summing compositing weights $T\alpha$ from Gaussians centered within each fixed body region or surface neighborhood, averaged over target core pixels.
Despite FM's improved background appearance, local contributions for the people and wall remain far below those for the shop sign (Figure~\ref{fig:mall_observation_representation}, right).
Most rendered opacity on the wall comes from Gaussians outside its defined neighborhood, so improved appearance can coexist with weak local contribution.
These final-model diagnostics describe the selected regions, whose sizes and temporal occupancy differ; full definitions appear in the supplement.

\subsection{Binaural Audio Reconstruction}
\label{sec:exp_audio}

\begin{table*}[!t]
    \centering
    \caption{\textbf{Concert visual editing.}
    Four fixed views, 0--10\,s, $1280\!\times\!720$; unedited controls share edited GT.
    Mask PSNR: target silhouette or relocation union, excluding shadows.
    ProPainter/ReVideo: removal/relocation, with ReVideo resized.}
    \vspace{-3mm}\label{tab:visual_editing}
    \footnotesize
    \setlength{\tabcolsep}{2pt}
    \renewcommand{\arraystretch}{1.08}
    \begin{tabular*}{\textwidth}{@{\extracolsep{\fill}}ll*{4}{c}@{\hspace{9pt}}*{4}{c}@{}}
        \toprule
        \multirow{2}{*}{Method} & \multirow{2}{*}{Output} & \multicolumn{4}{c}{Removal} & \multicolumn{4}{c}{Relocation} \\
        \cmidrule(lr){3-6}\cmidrule(lr){7-10}
        & & PSNR$\uparrow$ & SSIM$\uparrow$ & LPIPS$\downarrow$ & Mask PSNR$\uparrow$ & PSNR$\uparrow$ & SSIM$\uparrow$ & LPIPS$\downarrow$ & Mask PSNR$\uparrow$ \\
        \midrule
        Input video & Unedited & 23.182 & 0.8784 & 0.0684 & 8.391 & 22.879 & 0.8747 & 0.0763 & 8.597 \\
        ProPainter~\cite{zhou2023propainter} / ReVideo~\cite{mou2024revideo} & Edited & 25.405 & 0.8848 & 0.0599 & 12.196 & 25.252 & 0.7369 & 0.1287 & 18.505 \\
        \midrule
        \multirow{2}{*}{FTGS++~\cite{lee2026freetimegspp}} & Unedited & 22.474 & 0.8018 & 0.1692 & 8.554 & 22.237 & 0.7984 & 0.1759 & 8.763 \\
        & Edited & 23.727 & 0.8061 & 0.1649 & 11.151 & 23.566 & 0.8038 & 0.1669 & 11.444 \\
        \cmidrule(l){1-10}
        \multirow{2}{*}{TRASE~\cite{li2024trase}} & Unedited & 22.448 & 0.8209 & 0.1301 & 8.474 & 22.189 & 0.8173 & 0.1371 & 8.670 \\
        & Edited & 25.482 & 0.8373 & 0.1042 & 17.787 & 25.311 & 0.8350 & 0.1058 & 17.770 \\
        \bottomrule
    \end{tabular*}
    \vspace{-5mm}
\end{table*}

Across eight scenes, we predict target-view binaural audio from a synchronized spatial mixture recorded at another view.
Models are trained separately per scene on 46 fixed views and evaluated on four held-out fixed views, with shared preprocessing and scoring for indoor and outdoor scenes.
Methods share one input training-view recording per scene; full settings appear in the supplement.

We compare AudioGS~\cite{bi2026audiogs}, ViGAS~\cite{chen2023nvas}, and AV-Cloud~\cite{chen2024avcloud} (Full), with binaural copying as a reference baseline.
AudioGS uses audio and listener poses, while ViGAS and AV-Cloud additionally use visual observations.
Inputs exclude dry sources, test-view images, and engine geometry or acoustic responses.

Table~\ref{tab:audio_observed} averages 480 one-second test windows per scene. STFT is complex-spectrum L2 error; RTE is an auxiliary discrepancy between $RT_{60}$ estimates from predicted and reference audio~\cite{chen2023nvas}.
ViGAS achieves the lowest MAG and ENV in all eight scenes and the lowest STFT error in seven, showing consistent spectral and envelope gains indoors and outdoors.
Rankings differ for other attributes: AV-Cloud leads interaural log-energy ratio (LRE) in five scenes, whereas AudioGS or binaural copying leads DPAM in six.
DPAM~\cite{manocha2020dpam} and CDPAM~\cite{manocha2021cdpam} score mono-folded perceptual similarity; LRE measures interaural balance.

\subsection{Audiovisual Editing}
\label{sec:exp_editing}

Our Concert case study examines target editing and geometry-dependent acoustic changes.
Visual references rerender the scene with the violinist, violin, and bow removed or shifted horizontally by 1.5\,m, keeping other actors and the timeline fixed.
Audio removal is scored against the original piano stem; a separate panel intervention recomputes propagation.

\noindent\textbf{Setup.}
ProPainter~\cite{zhou2023propainter} and ReVideo~\cite{mou2024revideo} edit observed videos at four evaluation views; FTGS++ deletion/translation and TRASE~\cite{li2024trase} render these views from 46 fixed training cameras.
Removal uses RGB-derived SAM~\cite{kirillov2023segment} masks, shared across methods at their respective input views; engine labels are evaluation-only.
Each method has an unedited control; full-frame, target-mask, and complementary-region scores all use edited GT.

\begin{figure*}[!t]
    \centering
    \includegraphics[width=\textwidth]{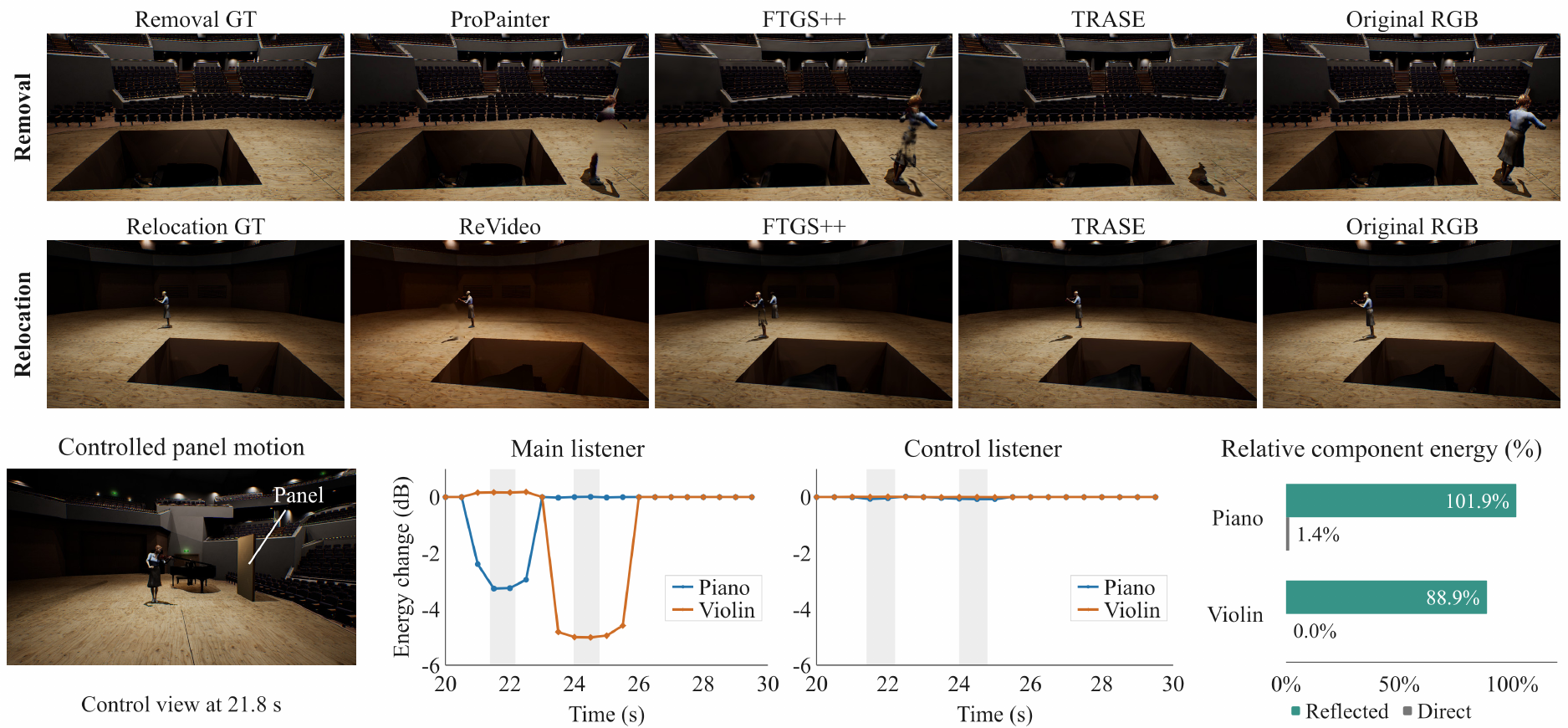}
    \vspace{-7mm}\caption{\textbf{Editing and controlled occlusion in Concert.}
    Top: time-matched removal (rear) and relocation (front oblique), comparing edited GT, video editors, FTGS++, TRASE, and original RGB.
    Bottom: moving panel, total unit-response energy changes at two listeners relative to a parked panel (shared scale), and relative direct/reflected energy at the main listener (100\%: each component with the parked panel).
    Shading: piano/violin holds; points: paired 2\,Hz responses.}
    \vspace{-5mm}\label{fig:av_editing}
\end{figure*}

\noindent\textbf{Visual removal.}
All editors improve over their unedited controls (Table~\ref{tab:visual_editing}).
ProPainter scores best on full-frame SSIM and LPIPS, while TRASE has the highest Mask PSNR.
Figure~\ref{fig:av_editing} (first row) shows target fragments left by ProPainter and FTGS++; TRASE removes the performer more completely but retains its shadow.

\Needspace{3\baselineskip}
\noindent\textbf{Visual relocation.}
ReVideo achieves the highest Mask PSNR, but TRASE has better SSIM and LPIPS at similar full-frame PSNR (Table~\ref{tab:visual_editing}).
Figure~\ref{fig:av_editing} (second row) shows duplicated content in FTGS++ and changes to the performer's appearance and background in ReVideo.
Relative to their respective unedited outputs, ReVideo loses 1.91\,dB of PSNR outside the target union, whereas FTGS++ and TRASE change by less than 0.2\,dB (supplement).

\begin{table}[!tb]
    \centering
    \caption{\textbf{Concert audio removal.}
    Four-view means against the original piano stem.
    Unchanged/Silence: input mixture/zero output; Silence has no interaural scores.
    Lower is better; LRE/ILD: dB, IPD: rad.}
    \vspace{-3mm}\label{tab:audio_editing}
    \footnotesize
    \setlength{\tabcolsep}{1pt}
    \renewcommand{\arraystretch}{1.08}
    \begin{tabular*}{\linewidth}{@{\extracolsep{\fill}}clccccc@{}}
        \toprule
        Time & Method & MAG$\downarrow$ & LRE$\downarrow$ & MR-STFT$\downarrow$ & ILD$\downarrow$ & IPD$\downarrow$ \\
        \midrule
        \multirow{4}{*}{\rotatebox[origin=c]{90}{0--10\,s}} & Unchanged & 0.10090 & 1.30209 & 6.78894 & 1.62532 & 0.13382 \\
        & SmartDJ~\cite{lan2026smartdj} & 0.07734 & 3.85581 & 5.40598 & 6.54081 & 1.41593 \\
        & Banquet~\cite{watcharasupat2024banquet} & 0.10072 & 1.30032 & 6.78272 & 1.62508 & 0.13381 \\
        & Silence & 0.01296 & -- & 1.05214 & -- & -- \\
        \midrule
        \multirow{4}{*}{\rotatebox[origin=c]{90}{20--30\,s}} & Unchanged & 0.06298 & 0.21442 & 1.30935 & 0.44705 & 0.05603 \\
        & SmartDJ~\cite{lan2026smartdj} & 0.18376 & 3.56959 & 1.60936 & 5.53048 & 1.36872 \\
        & Banquet~\cite{watcharasupat2024banquet} & 0.06297 & 0.21449 & 1.30920 & 0.44736 & 0.05605 \\
        & Silence & 0.23699 & -- & 2.57097 & -- & -- \\
        \bottomrule
    \end{tabular*}
    \vspace{-5mm}
\end{table}

\noindent\textbf{Audio removal.}
We compare text-guided SmartDJ~\cite{lan2026smartdj} with dry-query-conditioned Banquet~\cite{watcharasupat2024banquet}.
MR-STFT measures multiresolution sum/difference spectral errors; ILD and IPD measure interaural level and phase errors.
Silence minimizes MAG and MR-STFT at 0--10\,s, where piano enters late, but maximizes both at 20--30\,s, when the instruments overlap (Table~\ref{tab:audio_editing}).
SmartDJ lowers early spectral errors but worsens interaural errors; during overlap all five errors increase relative to the input mixture, while Banquet stays nearly unchanged.
Together, these references distinguish target-edit accuracy from fidelity to surrounding imagery and retained sound.

\noindent\textbf{Controlled dynamic occlusion.}
We render paired video and recompute direct and reflected sound in Concert over 20--30\,s for a parked or moving panel, fixing dry emissions, performer motion, and two camera--listener poses.
The panel sequentially crosses the piano and violin paths to the main listener; the control listener retains direct visibility.
Figure~\ref{fig:av_editing} (bottom) measures energy in each source's response to an identical unit impulse, separating propagation changes from musical dynamics.
During the respective holds, total response energy drops by 3.26 and 5.00\,dB at the main listener, while control-listener changes stay below 0.07\,dB.
Direct sound is strongly attenuated or blocked, while reflected energy remains.
This case connects visible motion with listener-dependent acoustic responses; statistics and clips are in the supplement.

\section{Conclusion}
\label{sec:conclusion}

SynAVID combines minute-scale dynamics, fixed and moving multiview video, and synchronized binaural audio in a controllable synthetic dataset.
Mixed-observation configurations generally improve visual reconstruction, with persistent local failures; audio-method rankings differ across spectral, interaural, and perceptual metrics.
Geometric and instance annotations connect these observations to regional support and final representations, while paired editing references assess target changes and scene fidelity.
A controlled panel intervention further links geometry changes to listener-dependent acoustic responses.
SynAVID provides a resource for developing immersive audiovisual models; broader scene coverage and transfer to real data are directions for future work.

\clearpage
\section*{LLM Usage Statement}
We used OpenAI ChatGPT and Codex to support manuscript writing and editing, related-work and citation preparation, experiment design and result analysis, figure and diagram preparation, and dataset construction and annotation.
For these activities, authors reviewed, edited, or approved tool-generated outputs through interactive use (\emph{Generate}).
For code and implementation, Codex also carried out multi-step tasks autonomously with limited step-by-step human direction (\emph{Agent}).
For research ideas, hypotheses, claims, and conclusions, AI tools assisted in editing or refining author-produced content (\emph{Assist}).
Major parts of the manuscript were also provided to OpenAI (ChatGPT/GPT) for reviewer-style feedback.
The authors take full responsibility for all submitted content, regardless of any LLM or AI-tool use.

\clearpage
\appendix
\backrefsetup{disable}
\setcounter{figure}{0}
\setcounter{table}{0}
\setcounter{equation}{0}
\renewcommand{\thefigure}{S\arabic{figure}}
\renewcommand{\thetable}{S\arabic{table}}
\renewcommand{\theequation}{S\arabic{equation}}
\renewcommand{\theHfigure}{supp.\arabic{figure}}
\renewcommand{\theHtable}{supp.\arabic{table}}
\renewcommand{\theHequation}{supp.\arabic{equation}}
\newcommand{\suppproject}[2]{\href{https://avery-xu.github.io/synavid-project/#1}{#2}}

\twocolumn[{%
  \renewcommand\twocolumn[1][]{#1}%
  \pdfbookmark[0]{Supplementary Material}{synavid-supplement}%
  {\maketitlesupplementary}%
  {\centering\small\textbf{Audiovisual examples.} See the \suppproject{}{project page}.\par}\medskip
}]
\section{Scene Inventory}
\label{sec:supp-scenes}

Table~\ref{tab:supp-scene-inventory} describes one 120-second episode for each of the eight scenes. All views share the same episode timeline. The source counts are therefore shared across views, while the binaural recording depends on each listener's position and orientation.

\begin{table*}[!t]
\caption{\textbf{Activities, sound sources, and references by scene.}
The upper four rows show indoor scenes and the lower four show outdoor scenes. The two count columns give episode totals (emitters and distinct dry signals) and concurrent emitters (median and peak over frame times). Snow records paw-contact onsets only, so the concurrent count is unavailable.}
\label{tab:supp-scene-inventory}
\centering
\fontsize{9.5}{11}\selectfont
\setlength{\tabcolsep}{2.5pt}
\renewcommand{\arraystretch}{1.05}
\newcommand{\inventoryhead}[2][l]{\raisebox{-.5\height}{\bfseries\shortstack[#1]{#2}}}
\begin{tabularx}{\linewidth}{@{}l >{\raggedright\arraybackslash}X >{\raggedright\arraybackslash}X cc >{\raggedright\arraybackslash}p{0.25\linewidth}@{}}
\toprule
\inventoryhead{Scene} & \inventoryhead{Activity} & \inventoryhead{Sound sources} & \inventoryhead[c]{Emitters /\\dry signals} & \inventoryhead[c]{Scheduled\\median/peak} & \inventoryhead{Example annotations and\\references} \\
\midrule
Concert & Pianist and violinist perform on moving platforms. The violinist descends the stage steps. & Piano and violin & 2/2 & 2/2 & Paired visual edits of the violinist and a moving-panel occlusion sequence. \\
Apartment & A resident dances and moves furniture. A dog explores as a robot vacuum cleans. & TV speakers, record player, dog, robot vacuum, heel strikes & 7/6 & 3/5 & 197 foot contacts with 3D position, foot side, and heel or forefoot label. \\
Gym & Athletes lift weights, pass a medicine ball, run on a treadmill, and cycle. & Music, speech, breathing, footfalls, weight and ball impacts, exercise machines & 40/38 & 7/16 & 30\,Hz 3D tracks of the pelvis, head, mouth, and both feet for ten people. \\
Mall & Shoppers ride escalators, walk between landings, and talk while standing or seated. & Conversation, music over loudspeakers, TV advertisements & 38/28 & 12/19 & Pedestrian masks and 3D regions for pedestrians, a shop sign, and a wall. \\
\midrule
Street & Pedestrians cross among passing vehicles or talk by the roadside. & Engines, tires, horns, footsteps, speech & 34/34 & 13/17 & 493 footfalls with frame, foot side, 3D position, and ground surface ID. \\
Rural & Deer, foxes, and pigs move near a passing vehicle. Crows take off and perch. & Animal calls, wingbeats, hoof and paw contacts, engine and tire noise & 19/19 & 5/10 & 128 hoof and 94 paw contacts with time and bone ID. Perches for six crow landings. \\
Park & People dance, practice tai chi, skateboard, and walk dogs. & Music, speech, footfalls, barking, skateboard wheels, fountain & 92/91 & 10/27 & 3D contact points for 4,124 footfalls and three skateboard placements. \\
Snow & Wolves cross the snow, crows fly and perch, and an SUV passes. & Wolf and crow calls, wingbeats, paw contacts, engine and tire noise & 39/39 & --- & 312 wolf paw contacts with paw ID, frame, 3D position, and snow surface normal. \\
\bottomrule
\end{tabularx}
\end{table*}

The first count column counts emitters in the scene and distinct dry signals. In Mall, six loudspeakers play the same music from different positions, contributing six emitters and one dry signal to the episode totals. The next column describes how many emitters are scheduled at the same time. At each frame we count an emitter once if any of its events is in progress, including pauses within the signal. The Mall values of 12 and 19 are the median and peak of that count over the episode. How prominently a listener hears any one of those emitters also depends on its signal level and the propagation path.

The final column gives examples of the annotations available for individual scenes. In Mall, pedestrian masks and scene geometry let us measure when each training camera saw a pedestrian at a particular location, and compare that support with a shop sign and a corridor wall (Section~\ref{sec:supp-visual}). In Concert, paired renders show how the image changes when the violinist is removed or relocated. For audio removal, Section~\ref{sec:supp-editing} uses each listener's piano stem under the original propagation conditions as the reference.

The base recordings pair each RGB sequence with its mixed binaural audio, camera and listener parameters, timestamps, and split assignment. The dry signals, emission schedules, and bindings to scene emitters are shared across views. The \suppproject{}{project page} includes a one-minute audiovisual tour of all eight scenes, the annotated Mall regions, and the Concert editing and occlusion comparisons. Additional viewpoints, depth maps, surface labels, and geometry can be exported on request for a specified episode.

\section{Audiovisual Acquisition}
\label{sec:supp-acquisition}

\noindent\textbf{A frame and its binaural recording.}
For frame $f$, we evaluate the actors, objects, and sound emitters at episode time $t_f=f/30$ seconds. Rendering that state from each camera's pose and intrinsics gives corresponding views even when the cameras are rendered in separate passes. Each listener follows its camera's position, forward direction, and up direction. In Concert, for example, a point at Unreal coordinates $(x,y,z)$ in centimeters has acoustic coordinates $(x,z,y)/100$ in meters. At 30\,fps and 48\,kHz, one frame interval spans 1,600 audio samples per ear. This clock aligns the image with the sound \emph{received} during that interval. A contact visible in the image can be heard later because sound takes time to reach the listener.

The native RGB frames are $3840\times2160$. Each moving camera keeps a single identifier throughout its trajectory and has a pose for every frame. Concert was rendered with motion blur and depth of field disabled. For visual reconstruction (Section~\ref{sec:supp-visual}), we scale both image dimensions and the camera intrinsics by one sixth to preserve the field of view.

\noindent\textbf{Comparing two listeners at the Concert stage lift.}
In Figure~3 of the main paper, listener A stays at a fixed camera (\texttt{cam45}) and listener B follows a moving immersive camera (\texttt{immersive02}) on the other side of the stage. We chose the pair from their poses and images before inspecting the audio levels. The two images show the same instant at 45\,s. The sound traces cover the same eight seconds, from 41 to 49\,s, for both listeners. To compare how a piano passage reaches the two positions, follow the piano row across A and B at the same episode time and compare the same ear. Each envelope gives the root-mean-square amplitude of the received instrument signal in successive 20\,ms windows. The envelope for each ear is mirrored about its own centerline. Read its height above the line on the same linear amplitude scale for all eight traces. The four \suppproject{\#listening}{binaural examples on the project page} preserve these relative levels.

The \emph{Blocked} traces describe the geometry of the direct path at each animation frame. For the piano, we trace lines from 128 points on the soundboard to the listener's center. A count of 0 means that all points have a clear path, 128 means that all paths are obstructed, and intermediate values indicate partial visibility. Separately, a frequency-dependent radiation pattern derived from grand-piano measurements~\cite{brandner2020piano} determines how the instrument radiates in different directions. The violin is modeled as an omnidirectional point source with a single ray to the listener's center. Its trace takes only two values, 0 for clear and 1 for blocked. Each instrument thus has one visibility trace per listener and a received envelope for each ear.

During the highlighted interval, all 128 piano paths are blocked at A in 67 of the 240 frames. At B, all 128 are clear throughout the same interval. Over the full eight-second excerpt, the piano has a higher RMS amplitude at A than at B in both ears. The envelopes include reflected sound and retain the effects of propagation distance and source radiation. For the violin, the ray stays clear at A and is blocked over two intervals at B.

The middle panel places these observations on the full camera paths. Summing distances between consecutive camera centers over 120\,s gives 27.84\,m for the environment trajectory, 42.04\,m for the performer-following trajectory, and 28.82\,m for the immersive trajectory. Listener B travels 3.56\,m in the highlighted eight seconds. Between consecutive half-second reflection updates in this interval, the largest displacements are 0.26\,m for B, 0.32\,m for the piano emitter, and 0.38\,m for the violin emitter.

\noindent\textbf{Geometry and materials for sound propagation.}
Rigid objects follow their scene transforms. For articulated bodies, simplified shapes attached to the bones approximate the volume occupied by the moving body. We use asset identity and rendered views to assign frequency-dependent coefficients from reference materials~\cite{liu2025acousticrooms,valveSteamAudio}. In Apartment, 24 single-glazed panes close the four window openings in the acoustic model, while the bedroom doorway remains open. The propagation solver uses these surfaces to determine which paths meet glass and which pass through an opening. RGB images and target silhouettes are rendered from the visual scene geometry.

\noindent\textbf{Reflection updates and response duration.}
Steam Audio~\cite{valveSteamAudio} evaluates propagation at successive scene states. The \emph{update rate} specifies how often the solver samples the changing sources, listeners, and geometry. Across the eight episodes, direct paths update at 30\,Hz and reflections at a base rate of 2\,Hz, or once every 15 video frames. Rural adds reflection states around sound onsets. The Concert and Apartment recordings also include transmission through obstacles and paths computed with acoustic probes. Section~\ref{sec:supp-panel} gives the propagation settings for the separate moving-panel experiment.

The \emph{response duration} in Table~\ref{tab:supp-response-horizons} specifies how far after a single impulse we retain its modeled reflections at one sampled state. In Concert, two consecutive states can be sampled at 45.0 and 45.5\,s. Each has an eight-second reflection response. Sound emitted around 45.0\,s can thus continue to reverberate as sound emitted around 45.5\,s begins to arrive. The 0.5\,s interval controls how closely the simulation follows scene motion, while the eight-second duration controls how much of each reverberant tail enters the recording. Section~\ref{sec:supp-reflection-rate} varies the update rate while keeping this eight-second duration fixed.

\begin{table}[t]
  \centering
  \caption{\textbf{Duration of the reflected response to one impulse at a sampled scene state.} The values give the length retained in each scene's simulation. Reflections are updated at a base interval of 0.5\,s across scenes. In Street, the stored interval begins at the direct arrival.}
  \label{tab:supp-response-horizons}
  \small
  \begin{tabular*}{\columnwidth}{@{\extracolsep{\fill}}lrlr@{}}
    \toprule
    \textbf{Indoor} & \textbf{Seconds} & \textbf{Outdoor} & \textbf{Seconds} \\
    \midrule
    Concert   & 8 & Street & 0.5 \\
    Apartment & 1 & Rural  & 1   \\
    Gym       & 1 & Park   & 4.5 \\
    Mall      & 6 & Snow   & 6   \\
    \bottomrule
  \end{tabular*}
\end{table}

% Queue the double-column overview for the page where the experiments begin.
\begin{table*}[t]
  \centering
  \caption{\textbf{Inputs and scoring references for visual reconstruction, binaural reconstruction, and Concert interventions} (Sections~\ref{sec:supp-visual}--\ref{sec:supp-editing}).}
  \label{tab:supp-observations}
  \small
  \begin{tabularx}{\textwidth}{@{}l >{\raggedright\arraybackslash}X >{\raggedright\arraybackslash}X@{}}
    \toprule
    Task & Available input & Scoring reference \\
    \midrule
    Visual reconstruction & Calibrated training RGB, fixed or fixed and moving & RGB from complete held-out fixed and moving trajectories \\
    Binaural reconstruction & One fixed training-view recording at test time, with pose and method-specific visual inputs & Synchronous binaural recordings at four other, held-out fixed positions \\
    Concert video editing & Original RGB from each test view, with a SAM mask or projected motion control & Rerender after removing or translating the performer, violin, and bow \\
    Concert 3D editing & 46 fixed training views and RGB-derived target masks & The same edited references at four held-out views \\
    Concert audio removal & Each listener's original mix and a text instruction or dry violin query & Piano received at that listener under the original propagation conditions \\
    Concert panel intervention & Same dry signals, source motion, and two listener poses & Paired simulations with the panel parked or moving \\
    \bottomrule
  \end{tabularx}
\end{table*}

We used pilot responses to choose the duration in several scenes. In Apartment and Gym, the pilots ran for 12\,s, with a 0.25\,s margin beyond the $-50$\,dB tail threshold. In Park, four source--listener pairs gave durations of 1.5--4.5\,s. The music loudspeaker in the square paired with the moving listener determined the 4.5\,s setting. In Street, four comparisons with a 6\,s reference put at most 0.149\% of native-band energy beyond the selected interval.

\noindent\textbf{Binaural synthesis and replay checks.}
In Eq.~(1) of the main paper, the emitted signal includes each source's event timing and gain envelope. We apply the time-varying propagation response to this signal and use head-related transfer functions to map arrival directions to the left and right ears. The final playback gain is shared across listeners in a scene.

Gym illustrates the synthesis settings in detail. Steam Audio 4.8.1 traces 65,536 rays with up to 16 bounces and 1,024 diffuse samples. It stores reflected energy in three bands centered at 400, 2,500, and 15,000\,Hz, with arrival direction represented by second-order spherical harmonics (ACN/N3D). We reconstruct directional waveforms from the energy fields and use ACN/SN3D for binaural decoding with the measured neutral-head FABIAN response~\cite{brinkmann2017fabian}. At a reflection update, we crossfade the input to successive responses over 1,920 samples (40\,ms), allowing earlier emissions to decay through their existing responses. The saved energy fields, carrier and filter settings, direct-path records, and dry signals support resynthesis.

We checked whether processing an interval in pieces changes its propagation records. In Gym, we processed frames 223--227 of \texttt{immersive01} for all 40 emitters, first in one interval and then with a split at frame 225. All 200 emitter--frame records for direct sound, transmission, and probe paths matched byte for byte. In a separate test at frame 30, the treadmill motor yielded identical reflection energies when the solver loaded the geometry afresh and when it reused the same geometry.

In Apartment, we checked the final listener waveform. Replaying camera 01 from 74 to 82\,s reproduced the original waveform checksum. A separate check of the openings found that 1,238 sampled rays intersected the added glass panes and that the bedroom doorway remained open.

\section{Visual Reconstruction and Regional Diagnostics}
\label{sec:supp-visual}

Table~\ref{tab:supp-observations} lists what each method observes and which reference we use to score its output in Sections~\ref{sec:supp-visual}--\ref{sec:supp-editing}. For example, a video editor receives the original recording from the view it edits. A 3D editor observes the performer through training cameras and renders the edit at held-out views.

\subsection{Full-sequence reconstruction}

For each of the eight episodes in Table~\ref{tab:supp-scene-inventory}, we fit one model per configuration to all 3,600 time steps (120\,s at 30\,fps) for 330,000 optimization steps. We use $640\times360$ RGB images for training and evaluation to keep full-sequence fitting tractable. The fixed-only configuration (F) uses 46 training cameras. The mixed configuration (FM) adds seven moving trajectories indoors and eight outdoors. We evaluate FTGS++ with both configurations and use FM for STG and 4C4D. We hold out complete trajectories throughout training and initialization. Each indoor episode has four fixed and three moving test views, and each outdoor episode has four fixed and two moving test views.

\noindent\textbf{Model inputs and initialization.}
Training samples the cameras uniformly and uses their RGB images, intrinsics, and poses. FTGS++~\cite{lee2026freetimegspp} triangulates initial 3D points from training views. To initialize motion, F backprojects a flow correspondence in the target frame using the point's source-camera depth. FM triangulates the target position from corresponding rays in multiple views. Thus the F/FM comparison includes both the additional views and a change in velocity initialization. Both Concert runs disable Gaussian relocation during training. STG~\cite{li2024spacetime} uses a single full-sequence model initialized from training-view keyframe points. Our 4C4D~\cite{zhou20264c4d} adapter supplies a training-derived point cloud in place of its default MASt3R initialization. Engine geometry and instance masks enter the Mall diagnostics below, after fitting.

\noindent\textbf{Scoring held-out views.}
For Table~3 of the main paper, we compare floating-point predictions clipped to $[0,1]$ with the corresponding JPEG reference frames decoded to RGB on the same scale. Each frame receives one PSNR score from its full-image RGB mean squared error, one SSIM score using an $11\times11$ Gaussian window ($\sigma=1.5$, same padding), and one LPIPS score using AlexNet v0.1 with inputs in $[-1,1]$. We average the 3,600 frame scores within each test view and then weight views equally within each fixed or moving group. The qualitative comparisons in Figure~4 of the main paper use frames decoded from the saved videos.

\subsection{What the training views cover in Mall}

\begin{figure}[htbp]
  \centering
  \includegraphics[width=\linewidth]{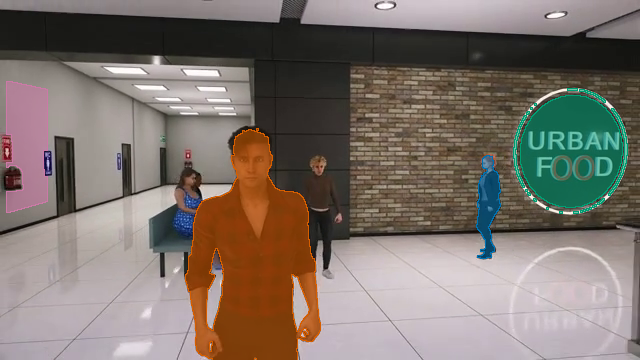}
  \caption{\textbf{Pixels used for the final Mall representation diagnostic.} At 68\,s, orange and blue identify the two pedestrians, green the shop sign, and purple the wall. The four mask interiors contain 19,694, 1,284, 8,989, and 4,160 pixels, respectively. The \suppproject{\#mall}{project page} shows the original frame beside the region overlay.}
  \label{fig:supp-mall-cores}
\end{figure}

Figure~\ref{fig:supp-mall-cores} identifies four regions in the held-out Mall view used in Figure~5 of the main paper (68\,s, frame 2040). The two pedestrians are difficult to recognize in the reconstructions. The \emph{URBAN FOOD} sign remains legible, and moving cameras supply most observations of the white corridor wall. We fix all four regions in world coordinates. For each pedestrian, we count visible pixels of that person only while they pass through the place they occupied at 68\,s. We then compare the training observations accumulated over 120\,s with the contribution of nearby Gaussians to the rendered image at 68\,s.

\noindent\textbf{Observations at a fixed place.}
For each pedestrian, we take the world-axis-aligned bounds of their deformed body at 68\,s and extend each side horizontally by 25\,cm and vertically by 10\,cm. Every 0.5\,s, native instance masks identify visible person pixels. When a body crosses its box boundary, we intersect each labeled pixel ray with the posed body mesh. The lower bound counts rays whose full candidate depth interval lies inside the box. The upper bound includes rays for which a target surface could lie inside it. We allow 2\,cm at the boundary. The two bounds give the range of support attributable to that location. For the two static surfaces, we count first-hit pixels on their respective world-space patches after masking visible pedestrians. This gives geometric visibility with pedestrian occlusion. Other moving occluders and view-dependent opacity can further change the visible pixels in RGB.

Let $n_R(c,t)$ be the resulting pixel count for region $R$ in camera $c$ at sampled time $t$. Each support curve averages $n_R$ over all cameras in its group at each time, including zero counts. Cumulative support sums over cameras and the 240 sample times, multiplying by 0.5\,s to give pixel-seconds. These totals retain differences in projected area as well as time in view. Each pedestrian contributes at three sampled instants, equivalent to 1.5 seconds in the support sum. The fixed array sees the sign throughout the episode and supplies 99.3\% of its support. Moving views supply 92.4\% of the wall's much smaller total (Table~\ref{tab:supp-mall-local}).

\noindent\textbf{Contribution to the final rendering.}
We render both final FTGS++ checkpoints at normalized time $2040/3599$. Over the interior pixels $p$ of a reference target mask $P_R$, we compute the mean contribution from Gaussians centered inside $R$ as
\begin{equation}
  L(R)=\frac{1}{|P_R|}\sum_{p\in P_R}
  \sum_{g:\,\mu_g(t)\in R} T_{g,p}\,\alpha_{g,p}.
\end{equation}
Here $\mu_g(t)$ is the center at the target time, $\alpha_{g,p}$ is its projected opacity, and $T_{g,p}$ accounts for all Gaussians in front of it. We use the same body boxes as above and a neighborhood extending 25\,cm on either side of each static surface patch. For the wall, FM raises mean rendered opacity over the 4,160 target pixels from 12.4\% to 92.3\%, yet centers in the wall neighborhood contribute only 0.10\% per pixel. At this view and time, most of the reconstructed wall opacity comes from Gaussians centered outside the wall neighborhood.

\begin{table}[t]
  \centering
  \caption{\textbf{Four regions in one Mall test view.} Support counts visible pixels in each region across all 53 training views, integrated over time. For each target pixel, we sum the compositing weights of Gaussians centered in the region. Local weight is the mean of these sums.}
  \label{tab:supp-mall-local}
  \small
  \begin{tabular*}{\columnwidth}{@{\extracolsep{\fill}}lrrr@{}}
    \toprule
    & 120\,s support & \multicolumn{2}{c@{}}{Local weight at 68\,s} \\
    \cmidrule(lr){3-4}
    Region & (pixel-seconds) & F (\%) & FM (\%) \\
    \midrule
    Man in red plaid & $1{,}130.5$--$1{,}612$ & 0.46 & 6.87 \\
    Woman in black   & $12{,}414.5$--$20{,}780.5$ & 1.27 & 5.35 \\
    Shop sign        & $9{,}051{,}281.5$ & 49.34 & 52.85 \\
    Corridor wall    & $18{,}896$ & 0.00 & 0.10 \\
    \bottomrule
  \end{tabular*}
\end{table}

\begin{figure*}[t]
  \centering
  \setkeys{Gin}{width=.211\textwidth}
  \newcommand{\mallpanel}[1]{\raisebox{-.5\height}{#1}}
  \begin{tabular}{@{}l@{\quad}c@{\hskip2pt}c@{\hskip6pt}c@{\hskip2pt}c@{}}
    & \multicolumn{2}{c}{Training camera-frames per surface point}
    & \multicolumn{2}{c}{Reconstruction error (RGB RMSE)} \\
    & F & FM & F & FM \\
    \shortstack[l]{Fixed view\\60\,s} &
    \mallpanel{\includegraphics{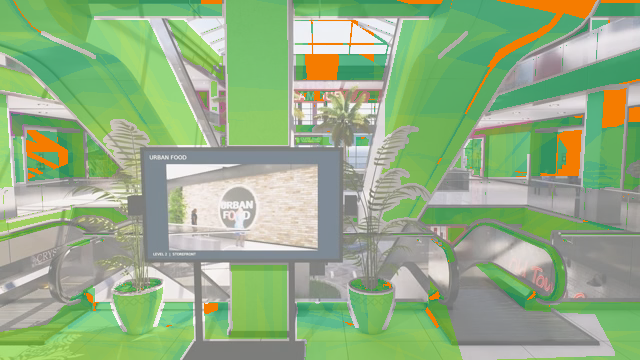}} &
    \mallpanel{\includegraphics{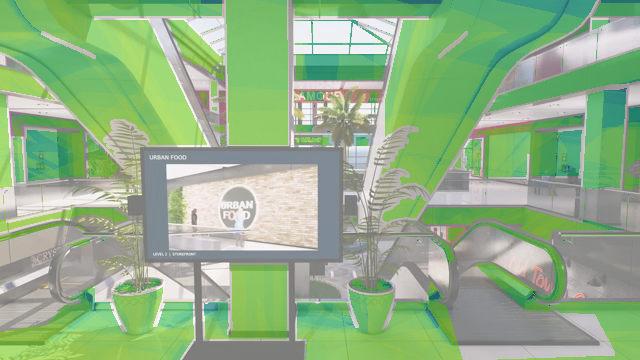}} &
    \mallpanel{\includegraphics{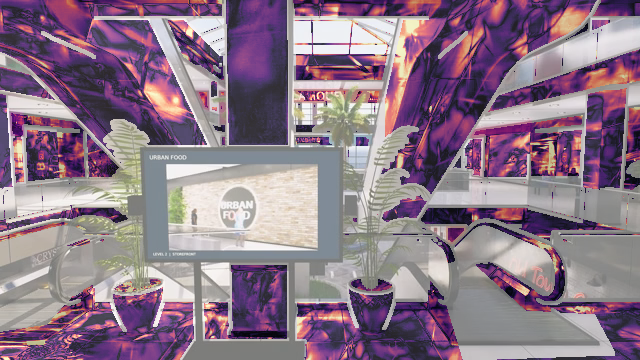}} &
    \mallpanel{\includegraphics{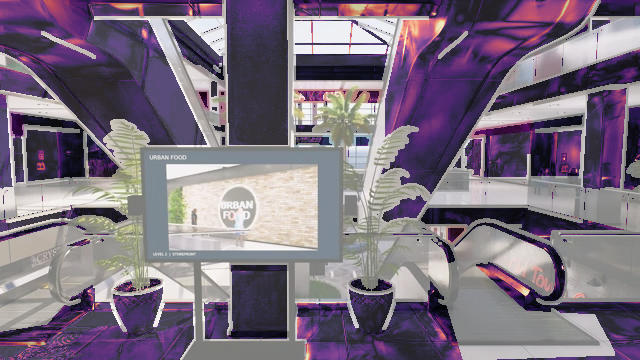}} \\[2pt]
    \shortstack[l]{Environment\\view, 69\,s} &
    \mallpanel{\includegraphics{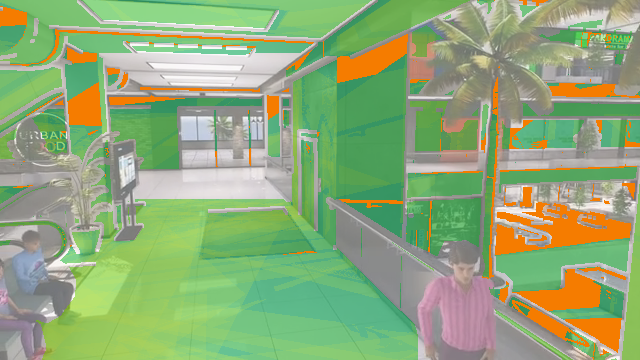}} &
    \mallpanel{\includegraphics{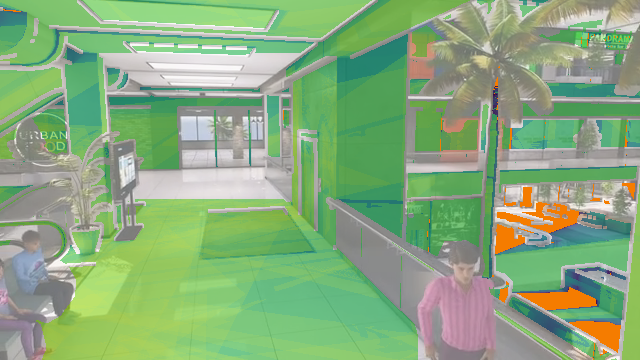}} &
    \mallpanel{\includegraphics{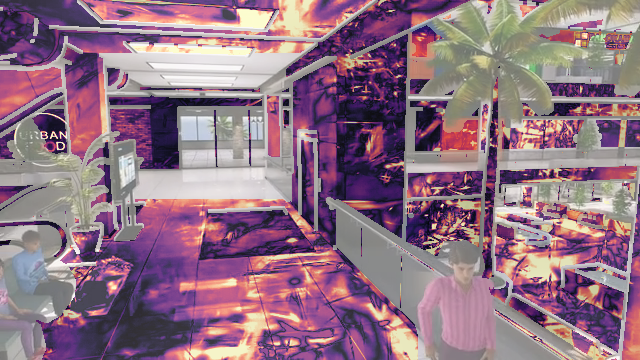}} &
    \mallpanel{\includegraphics{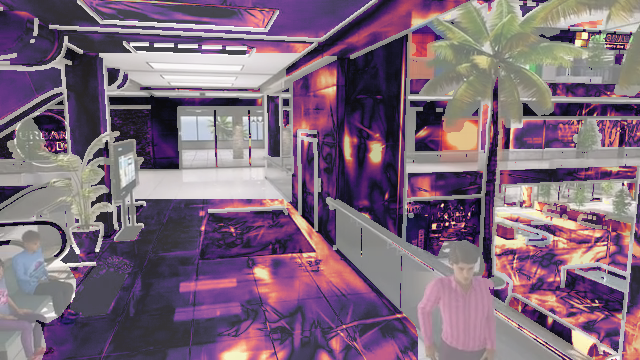}}
  \end{tabular}
  \caption{\textbf{Two additional views of Mall.} The coverage maps count training camera-frames with a geometric line of sight to each valid static test point. Orange marks zero observations. Purple through yellow spans 1 to 190,800 on a shared log scale. The error maps use a shared linear scale from black (0) to pale yellow (0.5 or higher). Faded RGB marks excluded pixels in all panels.}
  \label{fig:supp-mall-coverage}
\end{figure*}

\noindent\textbf{Coverage and image error in additional views.}
Figure~\ref{fig:supp-mall-coverage} extends the tracking-view example to a fixed view at 60\,s and an environment trajectory at 69\,s. As in the top row of Figure~5 of the main paper, we trace each valid test pixel to an opaque static surface and count training camera--frame pairs with an unobstructed line of sight to that point. Two cameras seeing the point at the same instant count twice. The maps use the static mesh and all 3,600 frame times. We exclude dynamic foreground, ambiguous materials, and depth boundaries in the test image. Points visible only from moving training cameras occupy 6.14\% and 9.17\% of valid pixels in the two views, respectively. On these pixels, FTGS++ PSNR rises from 12.97 to 16.75\,dB in the fixed view and from 11.84 to 15.03\,dB in the environment view. Pixels outside both training groups' coverage in the environment view show the opposite pattern (11.32 to 9.67\,dB). These regional scores pool RGB squared errors from the decoded H.264 predictions and references on identical masks. Table~3 of the main paper uses full-frame floating-point predictions.

\noindent\textbf{Depth of the Gaussians rendering the pedestrians.}
A separate check compares fixed-camera support over 68--70\,s and the full episode for three pedestrian boxes and three static surface patches. The boxes enclose the two pedestrians in Table~\ref{tab:supp-mall-local} and a third in the same test view. For each pedestrian, we examine the Gaussians contributing to pixels inside the silhouette at 68\,s. Across the three pedestrians and the two final FTGS++ models (F and FM), 79--99\% of the compositing weight on these interior pixels comes from Gaussians whose centers lie behind the corresponding body box as seen from the test camera.

\section{Binaural Audio Reconstruction}
\label{sec:supp-audio}

\noindent\textbf{Listening task.}
Given the mixture recorded at the ears of one listener, we predict what a listener at another fixed camera hears at the same instant. We train each method separately in each of the eight scenes using 46 fixed cameras. For each target camera and one-second training window, we draw the input listener uniformly from the other 45 cameras and pair their recordings of the same second. At test time, all methods use the same randomly chosen input listener for each scene (Table~\ref{tab:supp-input-listeners}). The four held-out listeners hear the same 120 one-second intervals, giving 480 target windows per scene. The binaural-copy baseline assigns the input recording's left and right channels directly to the corresponding target ears.

\begin{table}[t]
  \centering
  \caption{\textbf{Input recording for each scene in Table~4 of the main paper.} Camera names follow the episode metadata. The same input supplies all 120 test windows.}
  \label{tab:supp-input-listeners}
  \small
  \begin{tabular*}{\columnwidth}{@{\extracolsep{\fill}}llll@{}}
    \toprule
    Indoor & Input camera & Outdoor & Input camera \\
    \midrule
    Concert   & \texttt{cam07} & Street & \texttt{cam03} \\
    Apartment & \texttt{cam15} & Rural  & \texttt{fixed003} \\
    Gym       & \texttt{cam12} & Park   & \texttt{fixed49} \\
    Mall      & \texttt{mall\_fixed\_L2\_18} & Snow & \texttt{cam25} \\
    \bottomrule
  \end{tabular*}
\end{table}

\noindent\textbf{Inputs and synthesis.}
AudioGS~\cite{bi2026audiogs} uses the input binaural recording and the poses of the input and target listeners. For each ear, our adaptation predicts a spectral magnitude correction between $-6$ and $+6$\,dB and reconstructs the waveform using the phase from the corresponding input ear.

ViGAS~\cite{chen2023nvas} also sees the RGB frame at the midpoint of each one-second input recording (frame $30w+15$ for zero-indexed window $w$). A ResNet-18 encodes the frame. The synthesis network uses these features, the input binaural waveform, the input listener's pose, and the target's relative pose to predict the left and right waveforms directly.

AV-Cloud~\cite{chen2024avcloud} builds a visual representation once per scene. Using the first RGB frame from each of the 46 calibrated training cameras, we triangulate matching features into a sparse colored point cloud and cluster it into 256 anchors. For each audio window, the Full model averages the input recording's two channels to mono. Given this signal, the anchors, and the target listener's pose, it predicts a separate magnitude spectrum for each target ear and combines both with the phase of the mono input.

\noindent\textbf{Training and preprocessing.}
We resample each complete stereo recording from 48 to 16\,kHz, then split it into 120 non-overlapping one-second windows. Within each window, we apply a fifth-order Butterworth bandpass (150--7,999\,Hz) forward and backward to each ear. We preserve relative gains and delays and include quiet windows in training and evaluation. Each model produces 16,000 samples per ear, which we score before concatenating windows.

Each epoch covers all 5,520 target--window pairs in batches of 16. We use 50 training epochs for Table~4 of the main paper, except for AudioGS on Concert, which uses four epochs (1,380 optimizer steps). AudioGS uses Adam with an initial learning rate of 0.002, multiplied by 0.95 every five epochs. ViGAS uses Adam at 0.0003 with magnitude and waveform losses. AV-Cloud uses AdamW at 0.0005 with a log-magnitude loss.

\noindent\textbf{Scoring the predictions.}
For each metric in Table~4, we compare predictions with the synchronized target recording, average the 120 window scores for each listener, then average equally over the four listeners. Lower values indicate closer agreement for all eight metrics.

\emph{Waveform, envelope, and spectrum.} MSE averages squared waveform differences over samples and ears. Table~4 scales this value by $10^3$. ENV measures how amplitude varies over time by taking the Euclidean distance between the predicted and target Hilbert envelopes for each ear, then averaging the two distances. MAG and STFT use the same complex spectra, computed with a 512-point FFT, a 400-sample Hamming window, and a 160-sample hop with centered zero padding. MAG compares magnitudes, averaging absolute differences over time and frequency for each ear and summing over ears. STFT sums the two earwise Euclidean distances between complex spectra and is sensitive to both magnitude and phase.

\emph{Left--right energy balance.} LRE measures whether the prediction preserves the energy ratio between the two ears. For a stereo window $x$, let
\[
  r(x)=10\log_{10}\frac{\sum_n x_L[n]^2+10^{-5}}
                             {\sum_n x_R[n]^2+10^{-5}},
\]
LRE is the absolute difference $|r(\hat{x})-r(x)|$ in dB between the prediction $\hat{x}$ and the target $x$.

\emph{Estimated decay time.} RTE compares the $RT_{60}$ values that the speech-based estimator used in NVAS~\cite{chen2023nvas} assigns to the predicted and target recordings. We apply it to each ear of each one-second window and average the two absolute differences in seconds. For music, quiet passages, and outdoor sounds, the score describes agreement between the estimator's responses to those signals. Section~\ref{sec:supp-panel} examines propagation directly using impulse responses.

\emph{Perceptual similarity.} DPAM~\cite{manocha2020dpam} and CDPAM~\cite{manocha2021cdpam} compare the sound of the two recordings after mixing each to mono. We clip and quantize each ear to 16-bit PCM precision, average the channels, and resample to 22.05\,kHz. Each pretrained model scores the resulting pair of clips, target first.

\section{Editing and Dynamic Occlusion}
\label{sec:supp-editing}

We use three references to study edits in the Concert hall. A visual rerender shows how the scene looks after the performer is removed or relocated. The received piano stem gives the desired sound when the violin is silenced under the original propagation conditions. A separate pair of renders captures how moving an obstacle changes the propagation paths themselves.

\subsection{Removing and relocating a performer}

\noindent\textbf{Edited references and scoring masks.}
We replay frames 0--299 of Concert (ten seconds at 30\,fps). For removal we hide the violinist, violin, and bow. For relocation we translate all three by $(-1.5,0,0)$\,m in Unreal world coordinates, preserving their relative motion. Both rerenders retain the pianist, lighting, clock, and camera poses. The target mask for removal is the original visible silhouette of the violinist, violin, and bow. For relocation it is the union of their original and translated silhouettes. Both masks come from the engine and use zero dilation. The floor shadow disappears or moves in the edited RGB. It has no separate label, so changed shadow pixels beyond the silhouettes contribute to the outside-mask and full-frame scores.

The observed frames come from H.264 video and the edited references from lossless rerenders. We evaluate four views (\texttt{cam17}, \texttt{cam19}, \texttt{cam21}, and \texttt{cam23}). Figure~6 of the main paper shows frame 149 from the rear for removal and from a front-oblique view for relocation. The \suppproject{\#editing}{project page} shows both frames with the methods' outputs, the SAM input mask, and the engine scoring mask.

\noindent\textbf{Editing an observed video.}
ProPainter~\cite{zhou2023propainter} inpaints the target region of each original test-view video using RGB-derived masks from SAM ViT-H (checkpoint \nolinkurl{sam_vit_h_4b8939.pth})~\cite{kirillov2023segment}. Thin structures such as the bow can be missing from these input masks. ReVideo~\cite{mou2024revideo} receives the original view and an image-space motion control. To project the prescribed 1.50\,m world translation into a view, its control estimates depth from the first-frame mask height assuming a 1.70\,m performer. It then tracks the target in the source video. We process 22 independent 14-frame segments per view at 30\,fps. Removing the eight padding frames leaves 300 evaluation frames per view (1,200 in total). We resize the $1344\times768$ ReVideo outputs once with bicubic interpolation to the $1280\times720$ reference grid.

\noindent\textbf{Editing a reconstructed scene.}
FTGS++ and TRASE fit the same ten seconds from 46 fixed training cameras. For FTGS++~\cite{lee2026freetimegspp}, we fit 500,000 Gaussians for 30,000 steps, freeze the model, and select a target set by rendering into training views with SAM masks. The vote uses 93 mask observations from three training cameras. A supporting observation places at least half of a Gaussian's weight inside the mask and contributes at least 0.01 pixel-equivalents there. Selection requires two such observations, an in-mask share of at least 80\% across all observations, and base opacity of at least 0.01. This selects 17,205 Gaussians. We delete these Gaussians or translate their centers after evaluating their deformation. TRASE~\cite{li2024trase} uses its learned target association for both operations. Both 3D methods use RGB-derived SAM masks for target selection.

\noindent\textbf{Preservation outside the target.}
We pool squared RGB error over pixels and frames before converting it to PSNR. We average full-frame SSIM and AlexNet LPIPS over frames. Table~\ref{tab:supp-preservation} scores each method before and after editing against the same edited reference. On relocation, ReVideo improves the target-region score reported in the main paper, while PSNR outside the union drops by 1.91\,dB. For FTGS++ and TRASE, outside-mask PSNR changes by less than 0.2\,dB for either edit.

\begin{table}[t]
  \centering
  \caption{\textbf{Agreement with the edited reference outside the target silhouette.} PSNR (dB) pools pixels across four views, including surfaces where the edit changes shadows or reflections. Positive changes indicate closer agreement.}
  \label{tab:supp-preservation}
  \small
  \begin{tabular*}{\columnwidth}{@{\extracolsep{\fill}}llrrr@{}}
    \toprule
    Task & Method & Unedited & Edited & Change \\
    \midrule
    Removal & ProPainter & 27.80 & 27.98 & $+0.19$ \\
            & FTGS++    & 25.77 & 25.83 & $+0.07$ \\
            & TRASE     & 25.80 & 25.99 & $+0.18$ \\
    \midrule
    Relocation & ReVideo & 27.60 & 25.69 & $-1.91$ \\
               & FTGS++ & 25.67 & 25.74 & $+0.07$ \\
               & TRASE  & 25.68 & 25.87 & $+0.19$ \\
    \bottomrule
  \end{tabular*}
\end{table}

\subsection{Removing the violin from a binaural recording}

The two excerpts test different aspects of violin removal. At the four listeners, the piano first enters between 9.51 and 9.75\,s on the episode clock. The 0--10\,s excerpt therefore emphasizes suppression of the violin while the piano reference is mostly quiet. At 20--30\,s the instruments play together, so a successful edit must also preserve the piano. For each listener we take the corresponding ten seconds from the received piano stem, preserving its timing, gain, and reverberation from earlier notes under the original propagation conditions.

\noindent\textbf{Editing the mixture.}
Both editors take the original two-channel mixture and use pretrained weights. SmartDJ~\cite{lan2026smartdj} uses the instruction ``remove the sound of violin at the [direction]'' with right front, front, front, and left front for cameras 17, 19, 21, and 23, respectively. Its native 24\,kHz output is resampled to 48\,kHz. Banquet~\cite{watcharasupat2024banquet} takes the released dry violin excerpt from 20--30\,s as its query in both intervals. It estimates the violin from each observed mixture at its native 44.1\,kHz rate. We subtract the estimated stereo signal from the original mixture on the 48\,kHz grid. The ten-second input uses repeated reflection padding at the endpoints.

\noindent\textbf{Scoring and listening.}
For Table~6 of the main paper, we score each prediction against the received piano at the same listener. MAG and LRE are evaluated at 16\,kHz over the ten-second clip. At 48\,kHz, MR-STFT averages spectral convergence and log-magnitude errors over the channel sum and difference at three FFT sizes (512, 1,024, and 2,048). For interaural level difference (ILD), we average absolute level-ratio errors from 80\,Hz to 20\,kHz. For interaural phase difference (IPD), we average absolute wrapped phase errors from 80\,Hz to 1.5\,kHz. Both use a 1,536-sample Hann window, a 384-sample hop, and time--frequency bins whose mean reference magnitude across ears is within 40\,dB of its peak. The four cameras receive equal weight within each interval.

Table~6 gives spectral scores (MAG and MR-STFT) for the silent baseline, which has zero energy in both ears. Its MAG rises from 0.01296 in the first excerpt to 0.23699 when both instruments play.

To quantify how much Banquet subtracts, we divide the RMS of the estimated violin by that of the input mixture, pooling both ears and all ten seconds at each listener. Across the four listeners, the removed signal is 49.2--67.7\,dB below the mixture at 0--10\,s and 72.1--74.1\,dB below it at 20--30\,s. This small subtraction accounts for the close agreement between Banquet and the unchanged baseline in Table~6. The \suppproject{\#violin-removal}{listening examples on the project page} pair both edits with the input and piano reference at \texttt{cam17} for each interval.

\subsection{Changing the acoustic geometry}
\label{sec:supp-panel}

\noindent\textbf{Parked and moving panel.}
We render the 20--30\,s Concert excerpt twice for Figure~6 of the main paper. A wooden panel remains parked in one render and crosses the two instruments' direct paths in turn in the other. Both renders share the dry signals, performer motion, listener poses, ray seed, and pre-clip emission history. At the main listener, the panel obstructs the piano path 1.4--2.2\,s into playback, leaving 15 of the 128 soundboard samples visible during the hold. It then blocks the violin's single direct ray at 4.0--4.8\,s. Both direct paths stay open at the control listener. The parked panel can overlap the violinist in the image even when the line from the violin to the listener is clear.

\noindent\textbf{Measuring propagation with a unit impulse.}
A unit impulse gives each instrument the same excitation in the parked and moving conditions, independent of which note it plays. For each eight-second binaural response $h$, we compute $E(h)=\tfrac12\sum_{e,n} h_e[n]^2$ over ears $e$ and samples $n$. For total energy we first add the direct and reflected waveforms, retaining their signed cross term. For each hold, we average the two sampled response energies on a linear scale and report $10\log_{10}(\bar E_{\mathrm{moving}}/\bar E_{\mathrm{parked}})$. At the main listener, the violin's direct energy falls to zero during its hold, while its total response energy falls by 5.00\,dB because reflections still reach the listener. At the control listener the total changes by at most 0.063\,dB across both sources and holds (Table~\ref{tab:supp-panel}).

\noindent\textbf{Listening to the performance.}
The four synchronized clips play the actual music at the two listeners with the panel parked or moving. They preserve relative levels across the four clips and include the reverberant tails of earlier notes. In the \suppproject{\#occlusion}{player on the project page}, select \emph{Main}, \emph{Moving}, and \emph{Violin}, then switch between \emph{Direct} and \emph{Reflected} at 4.0--4.8\,s to hear what each path carries during the hold. The broadband curve shows the impulse-response measurement. The three-band plot shows the change in energy of the received music over each hold.

\noindent\textbf{Panel and propagation settings.}
The $1.20\times2.05$\,m panel is 8\,cm thick. Its wood absorption coefficients are 0.150, 0.130, and 0.130 at 400, 2,500, and 15,000\,Hz, with scattering 0.15. We recompute direct sound and reflections with transmission and probe pathing disabled. The renderer adds the geometric flight time $d/(340\,\mathrm{m\,s}^{-1})$ to the solver's excess delay and compensates the 16-sample decoder latency. Across the two sources and listeners, the geometric delays range from 8.60 to 22.44\,ms.

\begin{table}[t]
  \centering
  \caption{\textbf{Response-energy change with a moving panel (dB).} Absolute episode times are 21.5 and 22.0\,s for the piano hold, and 24.0 and 24.5\,s for the violin hold. A fully blocked violin ray gives zero direct energy in the moving condition.}
  \label{tab:supp-panel}
  \footnotesize
  \begin{tabular*}{\columnwidth}{@{\extracolsep{\fill}}lllrrr@{}}
    \toprule
    Hold & Listener & Source & Direct & Reflected & Total \\
    \midrule
    Piano & Main    & Piano  & $-18.622$ & $+0.080$ & $-3.258$ \\
          &         & Violin & $0$       & $+0.449$ & $+0.165$ \\
          & Control & Piano  & $0$       & $-0.099$ & $-0.050$ \\
          &         & Violin & $0$       & $+0.027$ & $+0.010$ \\
    \midrule
    Violin & Main    & Piano  & $0$       & $+0.011$ & $+0.005$ \\
           &         & Violin & $-\infty$ & $-0.513$ & $-5.003$ \\
           & Control & Piano  & $0$       & $-0.126$ & $-0.063$ \\
           &         & Violin & $0$       & $+0.002$ & $+0.001$ \\
    \bottomrule
  \end{tabular*}
\end{table}

\subsection{Reflection update rate and computational cost}
\label{sec:supp-reflection-rate}
Section~\ref{sec:supp-acquisition} describes the base reflection update rate of 2\,Hz used for the recordings. To measure the cost and accuracy of this choice during a change in visibility, we follow the violin at the main listener from 23 to 26\,s as the panel enters, holds, and leaves its direct path. We solve reflections at 2, 10, or 30\,Hz (every 15, 3, or 1 animation frames). The direct path updates at 30\,Hz throughout. All three settings use the same eight-second response duration, the production crossfade, and matching response snapshots at coincident times.

We emit a unit impulse at each test time and integrate only its reflected response over both ears. We compare the resulting energy $E_r$ at rate $r$ with the 30\,Hz reference using $10\log_{10}(E_r/E_{30})$. All reported percentiles use the same 60 emission times that fall between updates of both coarser grids. We time the corresponding three seconds of binaural violin music separately.

\begin{figure}[t]
  \centering
  \includegraphics[width=\columnwidth]{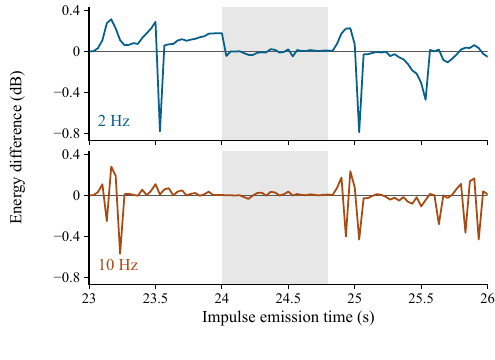}
  \caption{\textbf{Reflection updates as the panel crosses the violin's path.} Each curve gives the signed difference in reflected energy from the 30\,Hz reference for impulses emitted at successive animation frames. The shaded interval marks the panel's hold across the direct path.}
  \label{fig:supp-reflection-rate}
\end{figure}

Increasing the reflection update rate from 2 to 10\,Hz reduces the median absolute difference from 0.048 to 0.012\,dB and the maximum from 0.789 to 0.571\,dB (Figure~\ref{fig:supp-reflection-rate}). The 95th percentile rises from 0.320 to 0.404\,dB (Table~\ref{tab:supp-reflection-rate}). Of the 60 evaluated frames, four exceed 0.4\,dB at 10\,Hz, compared with three at 2\,Hz. All four occur one frame after a 10\,Hz update, within the 40\,ms window in which the renderer crossfades successive reflection responses. For context, repeating one fixed pose at 24\,s with three ray seeds gives a 0.067\,dB range in reflected energy.

\begin{table}[t]
  \centering
  \caption{\textbf{Reflection accuracy and cost for three seconds of Concert audio.} Absolute energy differences use the 60 shared off-grid frames. Solve time ends when the native responses reach disk. WAV time includes that solve and all subsequent synthesis through writing the binaural clip.}
  \label{tab:supp-reflection-rate}
  \small
  \begin{tabular*}{\columnwidth}{@{\extracolsep{\fill}}rrrrrr@{}}
    \toprule
    & & \multicolumn{2}{c}{Difference (dB)} & \multicolumn{2}{c}{Time (s)} \\
    Rate (Hz) & States & Median & P95 & Solve & WAV \\
    \midrule
    2  &  7 & 0.048 & 0.320 &  17.53 &  27.42 \\
    10 & 31 & 0.012 & 0.404 &  75.97 &  98.05 \\
    30 & 91 & ---   & ---   & 223.47 & 276.72 \\
    \bottomrule
  \end{tabular*}
\end{table}

For timing, each rate independently solves every state on one Intel Core i9-13900KS with one solver thread. We average two runs in opposite rate orders. All six runs use the same 19 pre-clip states, prepared once in 57.78\,s and timed separately. Relative to 2\,Hz, the three-second WAV takes $3.58\times$ as long at 10\,Hz and $10.09\times$ as long at 30\,Hz. For this crossing, 2\,Hz produces the three-second recording in 27.42\,s. Its median difference from the 30\,Hz reference is 0.048\,dB in integrated reflected energy, with larger deviations over individual frames as the panel moves.


{\small
\begin{thebibliography}{45}
\providecommand{\natexlab}[1]{#1}
\providecommand{\url}[1]{\texttt{#1}}
\expandafter\ifx\csname urlstyle\endcsname\relax
  \providecommand{\doi}[1]{doi: #1}\else
  \providecommand{\doi}{doi: \begingroup \urlstyle{rm}\Url}\fi

\bibitem[Bai et~al.(2025)Bai, Xia, Fu, Wang, Mu, Cao, Liu, Hu, Bai, Wan, and
  Zhang]{bai2025recammaster}
Jianhong Bai, Menghan Xia, Xiao Fu, Xintao Wang, Lianrui Mu, Jinwen Cao, Zuozhu
  Liu, Haoji Hu, Xiang Bai, Pengfei Wan, and Di Zhang.
\newblock {ReCamMaster}: Camera-controlled generative rendering from a single
  video.
\newblock In \emph{Proceedings of the IEEE/CVF International Conference on
  Computer Vision (ICCV)}, pages 14834--14844, 2025.

\bibitem[Berghi et~al.(2022)Berghi, Volino, and
  Jackson]{berghi2022tragictalkers}
Davide Berghi, Marco Volino, and Philip J.~B. Jackson.
\newblock {Tragic Talkers}: A shakespearean sound- and light-field dataset for
  audio-visual machine learning research.
\newblock In \emph{Proceedings of the 19th ACM SIGGRAPH European Conference on
  Visual Media Production}, 2022.

\bibitem[Bi et~al.(2026)Bi, Zhong, Xu, Song, and Cheng]{bi2026audiogs}
Chunhao Bi, Houqiang Zhong, Zhixin Xu, Li Song, and Zhengxue Cheng.
\newblock {AudioGS}: Spectrogram-based audio gaussian splatting for sound field
  reconstruction.
\newblock \emph{arXiv preprint arXiv:2604.08967}, 2026.

\bibitem[Brandner et~al.(2020)Brandner, Meyer-Kahlen, and
  Frank]{brandner2020piano}
Manuel Brandner, Nils Meyer-Kahlen, and Matthias Frank.
\newblock Directivity pattern measurement of a grand piano for augmented
  acoustic reality.
\newblock In \emph{Fortschritte der Akustik -- DAGA}, 2020.

\bibitem[Brinkmann et~al.(2017)Brinkmann, Lindau, Weinzierl, van~de Par,
  M{\"u}ller-Trapet, Opdam, and Vorl{\"a}nder]{brinkmann2017fabian}
Fabian Brinkmann, Alexander Lindau, Stefan Weinzierl, Steven van~de Par, Markus
  M{\"u}ller-Trapet, Rob Opdam, and Michael Vorl{\"a}nder.
\newblock A high resolution and full-spherical head-related transfer function
  database for different head-above-torso orientations.
\newblock \emph{Journal of the Audio Engineering Society}, 65\penalty0
  (10):\penalty0 841--848, 2017.

\bibitem[Broxton et~al.(2020)Broxton, Flynn, Overbeck, Erickson, Hedman,
  DuVall, Dourgarian, Busch, Whalen, and Debevec]{broxton2020immersive}
Michael Broxton, John Flynn, Ryan Overbeck, Daniel Erickson, Peter Hedman,
  Matthew DuVall, Jason Dourgarian, Jay Busch, Matt Whalen, and Paul Debevec.
\newblock Immersive light field video with a layered mesh representation.
\newblock \emph{ACM Transactions on Graphics}, 39\penalty0 (4):\penalty0
  86:1--86:15, 2020.

\bibitem[Chen et~al.(2022)Chen, Schissler, Garg, Kobernik, Clegg, Calamia,
  Batra, Robinson, and Grauman]{chen2022soundspaces2}
Changan Chen, Carl Schissler, Sanchit Garg, Philip Kobernik, Alexander Clegg,
  Paul Calamia, Dhruv Batra, Philip Robinson, and Kristen Grauman.
\newblock {SoundSpaces 2.0}: A simulation platform for visual-acoustic
  learning.
\newblock In \emph{Advances in Neural Information Processing Systems}, 2022.

\bibitem[Chen et~al.(2023)Chen, Richard, Shapovalov, Ithapu, Neverova, Grauman,
  and Vedaldi]{chen2023nvas}
Changan Chen, Alexander Richard, Roman Shapovalov, Vamsi~Krishna Ithapu,
  Natalia Neverova, Kristen Grauman, and Andrea Vedaldi.
\newblock Novel-view acoustic synthesis.
\newblock In \emph{Proceedings of the IEEE/CVF Conference on Computer Vision
  and Pattern Recognition (CVPR)}, pages 6409--6419, 2023.

\bibitem[Chen and Shlizerman(2024)]{chen2024avcloud}
Mingfei Chen and Eli Shlizerman.
\newblock {AV-Cloud}: Spatial audio rendering through audio-visual cloud
  splatting.
\newblock In \emph{Advances in Neural Information Processing Systems}, 2024.

\bibitem[Chen et~al.(2024)Chen, Gebru, Richardt, Kumar, Laney, Owens, and
  Richard]{chen2024raf}
Ziyang Chen, Israel~D. Gebru, Christian Richardt, Anurag Kumar, William Laney,
  Andrew Owens, and Alexander Richard.
\newblock Real acoustic fields: An audio-visual room acoustics dataset and
  benchmark.
\newblock In \emph{Proceedings of the IEEE/CVF Conference on Computer Vision
  and Pattern Recognition (CVPR)}, pages 21886--21896, 2024.

\bibitem[Donley et~al.(2021)Donley, Tourbabin, Lee, Broyles, Jiang, Shen,
  Pantic, Ithapu, and Mehra]{donley2021easycom}
Jacob Donley, Vladimir Tourbabin, Jung-Suk Lee, Mark Broyles, Hao Jiang, Jie
  Shen, Maja Pantic, Vamsi~Krishna Ithapu, and Ravish Mehra.
\newblock {EasyCom}: An augmented reality dataset to support algorithms for
  easy communication in noisy environments.
\newblock \emph{arXiv preprint arXiv:2107.04174}, 2021.

\bibitem[Gao and Grauman(2019)]{gao2019fairplay}
Ruohan Gao and Kristen Grauman.
\newblock {2.5D} visual sound.
\newblock In \emph{Proceedings of the IEEE/CVF Conference on Computer Vision
  and Pattern Recognition (CVPR)}, pages 324--333, 2019.

\bibitem[G{\"o}tz et~al.(2024)G{\"o}tz, G{\"o}tz, Meyer-Kahlen, Lee, Prawda,
  Habets, and Schlecht]{goetz2024mrtd}
Philipp G{\"o}tz, Georg G{\"o}tz, Nils Meyer-Kahlen, Kyung~Yun Lee, Karolina
  Prawda, Emanu{\"e}l A.~P. Habets, and Sebastian~J. Schlecht.
\newblock A multi-room transition dataset for blind estimation of energy decay.
\newblock In \emph{2024 18th International Workshop on Acoustic Signal
  Enhancement (IWAENC)}, pages 299--303. IEEE, 2024.

\bibitem[Grauman et~al.(2024)Grauman, Westbury, Torresani, Kitani, Malik,
  Afouras, Ashutosh, Baiyya, Bansal, Boote, Byrne, Chavis, Chen, Cheng, Chu,
  Crane, Dasgupta, Dong, Escobar, Forigua, Gebreselasie, Haresh, Huang, Islam,
  Jain, Khirodkar, Kukreja, Liang, Liu, Majumder, Mao, Martin, Mavroudi,
  Nagarajan, Ragusa, Ramakrishnan, Seminara, Somayazulu, Song, Su, Xue, Zhang,
  Zhang, Castillo, Chen, Fu, Furuta, Gonzalez, Gupta, Hu, Huang, Huang, Khoo,
  Kumar, Kuo, Lakhavani, Liu, Luo, Luo, Meredith, Miller, Oguntola, Pan, Peng,
  Pramanick, Ramazanova, Ryan, Shan, Somasundaram, Song, Southerland, Tateno,
  Wang, Wang, Yagi, Yan, Yang, Yu, Zha, Zhao, Zhao, Zhu, Zhuo, Arbelaez,
  Bertasius, Damen, Engel, Farinella, Furnari, Ghanem, Hoffman, Jawahar,
  Newcombe, Park, Rehg, Sato, Savva, Shi, Shou, and Wray]{grauman2024egoexo4d}
Kristen Grauman, Andrew Westbury, Lorenzo Torresani, Kris Kitani, Jitendra
  Malik, Triantafyllos Afouras, Kumar Ashutosh, Vijay Baiyya, Siddhant Bansal,
  Bikram Boote, Eugene Byrne, Zach Chavis, Joya Chen, Feng Cheng, Fu-Jen Chu,
  Sean Crane, Avijit Dasgupta, Jing Dong, Maria Escobar, Cristhian Forigua,
  Abrham Gebreselasie, Sanjay Haresh, Jing Huang, Md~Mohaiminul Islam, Suyog
  Jain, Rawal Khirodkar, Devansh Kukreja, Kevin~J Liang, Jia-Wei Liu, Sagnik
  Majumder, Yongsen Mao, Miguel Martin, Effrosyni Mavroudi, Tushar Nagarajan,
  Francesco Ragusa, Santhosh~Kumar Ramakrishnan, Luigi Seminara, Arjun
  Somayazulu, Yale Song, Shan Su, Zihui Xue, Edward Zhang, Jinxu Zhang, Angela
  Castillo, Changan Chen, Xinzhu Fu, Ryosuke Furuta, Cristina Gonzalez, Prince
  Gupta, Jiabo Hu, Yifei Huang, Yiming Huang, Weslie Khoo, Anush Kumar, Robert
  Kuo, Sach Lakhavani, Miao Liu, Mi Luo, Zhengyi Luo, Brighid Meredith, Austin
  Miller, Oluwatumininu Oguntola, Xiaqing Pan, Penny Peng, Shraman Pramanick,
  Merey Ramazanova, Fiona Ryan, Wei Shan, Kiran Somasundaram, Chenan Song,
  Audrey Southerland, Masatoshi Tateno, Huiyu Wang, Yuchen Wang, Takuma Yagi,
  Mingfei Yan, Xitong Yang, Zecheng Yu, Shengxin~Cindy Zha, Chen Zhao, Ziwei
  Zhao, Zhifan Zhu, Jeff Zhuo, Pablo Arbelaez, Gedas Bertasius, Dima Damen,
  Jakob Engel, Giovanni~Maria Farinella, Antonino Furnari, Bernard Ghanem, Judy
  Hoffman, C.~V. Jawahar, Richard Newcombe, Hyun~Soo Park, James~M. Rehg,
  Yoichi Sato, Manolis Savva, Jianbo Shi, Mike~Zheng Shou, and Michael Wray.
\newblock {Ego-Exo4D}: Understanding skilled human activity from first- and
  third-person perspectives.
\newblock In \emph{Proceedings of the IEEE/CVF Conference on Computer Vision
  and Pattern Recognition (CVPR)}, pages 19383--19400, 2024.

\bibitem[Guo et~al.(2025)Guo, Pan, Zhu, Hu, Zhang, Tang, Yang, Wang, Zhang,
  Wang, Chen, Xu, Xu, Fan, Chen, Yu, Huang, Wu, and Zhao]{guo2025mrsaudio}
Wenxiang Guo, Changhao Pan, Zhiyuan Zhu, Xintong Hu, Yu Zhang, Li Tang, Rui
  Yang, Han Wang, Zongbao Zhang, Yuhan Wang, Yixuan Chen, Hankun Xu, Ke Xu,
  Pengfei Fan, Zhetao Chen, Yanhao Yu, Qiange Huang, Fei Wu, and Zhou Zhao.
\newblock {MRSAudio}: A large-scale multimodal recorded spatial audio dataset
  with refined annotations.
\newblock In \emph{Advances in Neural Information Processing Systems}, 2025.

\bibitem[{International Telecommunication Union}(2020)]{itu2020immersiveav}
{International Telecommunication Union}.
\newblock Collection of usage scenarios and current statuses of advanced
  immersive audio-visual systems.
\newblock Report ITU-R BT.2420-1, ITU, 2020.

\bibitem[{ISO/IEC JTC 1/SC 29/AG 3}(2022)]{mpeg2022immersivevideo}
{ISO/IEC JTC 1/SC 29/AG 3}.
\newblock White paper on {MPEG} immersive video.
\newblock White paper N0059, ISO/IEC, 2022.

\bibitem[{ISO/IEC JTC 1/SC 29/AG 3}(2024)]{mpeg2024immersiveaudio}
{ISO/IEC JTC 1/SC 29/AG 3}.
\newblock White paper on {MPEG-I} immersive audio.
\newblock White paper N150, ISO/IEC, 2024.

\bibitem[Jiang et~al.(2026)Jiang, Lan, Luo, Deng, Lai, Sucar, Rupprecht, Laina,
  Larlus, Zheng, and Vedaldi]{jiang2026syn4d}
Zeren Jiang, Yushi Lan, Yihang Luo, Yufan Deng, Zihang Lai, Edgar Sucar,
  Christian Rupprecht, Iro Laina, Diane Larlus, Chuanxia Zheng, and Andrea
  Vedaldi.
\newblock {Syn4D}: A multiview synthetic {4D} dataset.
\newblock \emph{arXiv preprint arXiv:2605.05207}, 2026.

\bibitem[Jin et~al.(2026)Jin, Chen, Lin, and Gao]{jin2026sonoworld4d}
Derong Jin, Xiyi Chen, Ming~C. Lin, and Ruohan Gao.
\newblock {SonoWorld4D}: From one video to a {4D} audio-visual scene via
  diffusion synchronization.
\newblock \emph{arXiv preprint arXiv:2610.40640}, 2026.

\bibitem[Kirillov et~al.(2023)Kirillov, Mintun, Ravi, Mao, Rolland, Gustafson,
  Xiao, Whitehead, Berg, Lo, Doll{\'a}r, and Girshick]{kirillov2023segment}
Alexander Kirillov, Eric Mintun, Nikhila Ravi, Hanzi Mao, Chloe Rolland, Laura
  Gustafson, Tete Xiao, Spencer Whitehead, Alexander~C. Berg, Wan-Yen Lo, Piotr
  Doll{\'a}r, and Ross Girshick.
\newblock Segment anything.
\newblock In \emph{Proceedings of the IEEE/CVF International Conference on
  Computer Vision (ICCV)}, pages 4015--4026, 2023.

\bibitem[Lan et~al.(2026)Lan, Hao, and Zhao]{lan2026smartdj}
Zitong Lan, Yiduo Hao, and Mingmin Zhao.
\newblock {SmartDJ}: Declarative audio editing with audio language model.
\newblock In \emph{International Conference on Learning Representations}, 2026.

\bibitem[Lee et~al.(2026)Lee, Kim, Kim, Kim, and Park]{lee2026freetimegspp}
Lucas~Yunkyu Lee, Soonho Kim, Youngwook Kim, Sangmin Kim, and Jaesik Park.
\newblock {FreeTimeGS++}: Secrets of dynamic gaussian splatting and their
  principles.
\newblock \emph{arXiv preprint arXiv:2605.03337}, 2026.

\bibitem[Li et~al.(2022)Li, Slavcheva, Zollh{\"o}fer, Green, Lassner, Kim,
  Schmidt, Lovegrove, Goesele, Newcombe, and Lv]{li2022neural3dvideo}
Tianye Li, Mira Slavcheva, Michael Zollh{\"o}fer, Simon Green, Christoph
  Lassner, Changil Kim, Tanner Schmidt, Steven Lovegrove, Michael Goesele,
  Richard Newcombe, and Zhaoyang Lv.
\newblock Neural {3D} video synthesis from multi-view video.
\newblock In \emph{Proceedings of the IEEE/CVF Conference on Computer Vision
  and Pattern Recognition (CVPR)}, pages 5521--5531, 2022.

\bibitem[Li et~al.(2024{\natexlab{a}})Li, Gladkova, Xia, and
  Cremers]{li2024trase}
Yun-Jin Li, Mariia Gladkova, Yan Xia, and Daniel Cremers.
\newblock {TRASE}: Tracking-free 4d segmentation and editing.
\newblock \emph{arXiv preprint arXiv:2411.19290}, 2024{\natexlab{a}}.

\bibitem[Li et~al.(2024{\natexlab{b}})Li, Chen, Li, and Xu]{li2024spacetime}
Zhan Li, Zhang Chen, Zhong Li, and Yi Xu.
\newblock Spacetime gaussian feature splatting for real-time dynamic view
  synthesis.
\newblock In \emph{Proceedings of the IEEE/CVF Conference on Computer Vision
  and Pattern Recognition (CVPR)}, 2024{\natexlab{b}}.

\bibitem[Liang et~al.(2023)Liang, Huang, Tian, Kumar, and Xu]{liang2023avnerf}
Susan Liang, Chao Huang, Yapeng Tian, Anurag Kumar, and Chenliang Xu.
\newblock {AV-NeRF}: Learning neural fields for real-world audio-visual scene
  synthesis.
\newblock In \emph{Advances in Neural Information Processing Systems}, 2023.

\bibitem[Liu et~al.(2025)Liu, Kumar, Calamia, Amengual, Murdock, Ananthabhotla,
  Robinson, Shlizerman, Ithapu, and Gao]{liu2025acousticrooms}
Xiulong Liu, Anurag Kumar, Paul Calamia, Sebasti{\`a}~V. Amengual, Calvin
  Murdock, Ishwarya Ananthabhotla, Philip Robinson, Eli Shlizerman,
  Vamsi~Krishna Ithapu, and Ruohan Gao.
\newblock Hearing anywhere in any environment.
\newblock In \emph{Proceedings of the IEEE/CVF Conference on Computer Vision
  and Pattern Recognition (CVPR)}, pages 5732--5741, 2025.

\bibitem[Lu et~al.(2024)Lu, Zhou, Xing, Pokhariya, Dey, Shah, Mavidipalli, Hu,
  Comport, Chen, and Sridhar]{lu2024diva360}
Cheng-You Lu, Peisen Zhou, Angela Xing, Chandradeep Pokhariya, Arnab Dey,
  Ishaan~Nikhil Shah, Rugved Mavidipalli, Dylan Hu, Andrew~I. Comport, Kefan
  Chen, and Srinath Sridhar.
\newblock {DiVa-360}: The dynamic visual dataset for immersive neural fields.
\newblock In \emph{Proceedings of the IEEE/CVF Conference on Computer Vision
  and Pattern Recognition (CVPR)}, pages 22466--22476, 2024.

\bibitem[Manocha et~al.(2020)Manocha, Finkelstein, Zhang, Bryan, Mysore, and
  Jin]{manocha2020dpam}
Pranay Manocha, Adam Finkelstein, Richard Zhang, Nicholas~J. Bryan, Gautham~J.
  Mysore, and Zeyu Jin.
\newblock A differentiable perceptual audio metric learned from just noticeable
  differences.
\newblock In \emph{Proceedings of Interspeech}, pages 2852--2856, 2020.

\bibitem[Manocha et~al.(2021)Manocha, Jin, Zhang, and
  Finkelstein]{manocha2021cdpam}
Pranay Manocha, Zeyu Jin, Richard Zhang, and Adam Finkelstein.
\newblock {CDPAM}: Contrastive learning for perceptual audio similarity.
\newblock In \emph{Proceedings of the IEEE International Conference on
  Acoustics, Speech and Signal Processing (ICASSP)}, 2021.

\bibitem[Mou et~al.(2024)Mou, Cao, Wang, Zhang, Shan, and
  Zhang]{mou2024revideo}
Chong Mou, Mingdeng Cao, Xintao Wang, Zhaoyang Zhang, Ying Shan, and Jian
  Zhang.
\newblock {ReVideo}: Remake a video with motion and content control.
\newblock In \emph{Advances in Neural Information Processing Systems}, 2024.

\bibitem[Nazarczuk et~al.(2026)Nazarczuk, Tanay, Moreau, Zhang, and
  P{\'e}rez-Pellitero]{nazarczuk2025charge}
Michal Nazarczuk, Thomas Tanay, Arthur Moreau, Zhensong Zhang, and Eduardo
  P{\'e}rez-Pellitero.
\newblock {Charge}: A comprehensive novel view synthesis benchmark and dataset
  to bind them all.
\newblock In \emph{Proceedings of the IEEE/CVF Conference on Computer Vision
  and Pattern Recognition (CVPR)}, pages 15323--15333, 2026.

\bibitem[Park et~al.(2021)Park, Sinha, Hedman, Barron, Bouaziz, Goldman,
  Martin-Brualla, and Seitz]{park2021hypernerf}
Keunhong Park, Utkarsh Sinha, Peter Hedman, Jonathan~T. Barron, Sofien Bouaziz,
  Dan~B Goldman, Ricardo Martin-Brualla, and Steven~M. Seitz.
\newblock {HyperNeRF}: A higher-dimensional representation for topologically
  varying neural radiance fields.
\newblock \emph{ACM Transactions on Graphics}, 40\penalty0 (6), 2021.

\bibitem[Pumarola et~al.(2021)Pumarola, Corona, Pons-Moll, and
  Moreno-Noguer]{pumarola2021dnerf}
Albert Pumarola, Enric Corona, Gerard Pons-Moll, and Francesc Moreno-Noguer.
\newblock {D-NeRF}: Neural radiance fields for dynamic scenes.
\newblock In \emph{Proceedings of the IEEE/CVF Conference on Computer Vision
  and Pattern Recognition (CVPR)}, pages 10318--10327, 2021.

\bibitem[Roh et~al.(2026)Roh, Lee, Jeong, Lee, Lee, Park, and
  Park]{roh2026dusting}
Minhyeok Roh, Minseo Lee, Jun~Young Jeong, Byeonghyeon Lee, Lucas~Yunkyu Lee,
  Jaesik Park, and Eunbyung Park.
\newblock Dusting the phantom: Gaussian allocation for fading dynamic content.
\newblock \emph{arXiv preprint arXiv:2610.21302}, 2026.

\bibitem[Sabater et~al.(2017)Sabater, Boisson, Vandame, Kerbiriou, Babon, Hog,
  Gendrot, Langlois, Bureller, Schubert, and Allie]{sabater2017lightfield}
Neus Sabater, Guillaume Boisson, Benoit Vandame, Paul Kerbiriou, Frederic
  Babon, Matthieu Hog, Remy Gendrot, Tristan Langlois, Olivier Bureller, Arno
  Schubert, and Valerie Allie.
\newblock Dataset and pipeline for multi-view light-field video.
\newblock In \emph{Proceedings of the IEEE Conference on Computer Vision and
  Pattern Recognition (CVPR) Workshops}, pages 30--40, 2017.

\bibitem[Sandu et~al.(2026)Sandu, Pathak, and
  Soundararajan]{sandu2026temporally}
Vikram Sandu, Mayurdeep Pathak, and Rajiv Soundararajan.
\newblock Temporally aware densification for dynamic {3D} gaussian splatting.
\newblock \emph{arXiv preprint arXiv:2606.23212}, 2026.

\bibitem[Tsui et~al.(2026)Tsui, Tuan, Lai, and Wang]{tsui2026stady4d}
Hao-Tang Tsui, Yu-Rou Tuan, Ethan Lai, and Chen-Yu Wang.
\newblock {StaDy4D}: Towards complete {4D} static-dynamic reconstruction with
  {SIGMA}.
\newblock In \emph{Proceedings of the IEEE/CVF Conference on Computer Vision
  and Pattern Recognition (CVPR) Workshops}, pages 4697--4707, 2026.

\bibitem[{Valve Corporation}()]{valveSteamAudio}
{Valve Corporation}.
\newblock {Steam Audio}.
\newblock \url{https://valvesoftware.github.io/steam-audio/}.
\newblock Version 4.8.1. Accessed November 2026.

\bibitem[Watcharasupat and Lerch(2024)]{watcharasupat2024banquet}
Karn~N. Watcharasupat and Alexander Lerch.
\newblock A stem-agnostic single-decoder system for music source separation
  beyond four stems.
\newblock In \emph{Proceedings of the International Society for Music
  Information Retrieval Conference}, 2024.

\bibitem[Yang et~al.(2025)Yang, Pan, Wang, Wang, Lin, Li, Wen, Lin, Tao, and
  Yu]{yang2025imvid}
Zhengxian Yang, Shi Pan, Shengqi Wang, Haoxiang Wang, Li Lin, Guanjun Li,
  Zhengqi Wen, Borong Lin, Jianhua Tao, and Tao Yu.
\newblock {ImViD}: Immersive volumetric videos for enhanced {VR} engagement.
\newblock In \emph{Proceedings of the IEEE/CVF Conference on Computer Vision
  and Pattern Recognition (CVPR)}, pages 16554--16564, 2025.

\bibitem[Zerkowski and Velho(2026)]{zerkowski2026audioworldsim}
Luis~Vitor Zerkowski and Luiz Velho.
\newblock {AudioWorldSim}: Realistic binaural audio datasets for world models.
\newblock \emph{arXiv preprint arXiv:2608.21075}, 2026.

\bibitem[Zhou et~al.(2026)Zhou, Yang, Han, Zhang, Shi, Xu, and
  Liu]{zhou20264c4d}
Junsheng Zhou, Zhifan Yang, Liang Han, Wenyuan Zhang, Kanle Shi, Shenkun Xu,
  and Yu-Shen Liu.
\newblock {4C4D}: 4 camera 4d gaussian splatting.
\newblock In \emph{Proceedings of the IEEE/CVF Conference on Computer Vision
  and Pattern Recognition (CVPR)}, 2026.

\bibitem[Zhou et~al.(2023)Zhou, Li, Chan, and Loy]{zhou2023propainter}
Shangchen Zhou, Chongyi Li, Kelvin C.~K. Chan, and Chen~Change Loy.
\newblock {ProPainter}: Improving propagation and transformer for video
  inpainting.
\newblock In \emph{Proceedings of the IEEE/CVF International Conference on
  Computer Vision}, 2023.

\end{thebibliography}
}
\end{document}